\documentclass[final,3p,times]{elsarticle}

\usepackage{amssymb}
\usepackage{amsmath}
\usepackage{algorithm}
\usepackage{algpseudocode}
\usepackage{microtype}
\usepackage{booktabs}
\usepackage{graphicx}
\usepackage[hidelinks]{hyperref}

\makeatletter
\def\ps@pprintTitle{%
  \let\@oddhead\@empty
  \let\@evenhead\@empty
  \def\@oddfoot{\reset@font\hfil\thepage\hfil}%
  \let\@evenfoot\@oddfoot}
\makeatother

\begin{document}

\begin{frontmatter}

\title{Explaining spatial information flow in short-term traffic forecasting models using a gated graph attention network}

\author[1]{Yue Li}
\author[1]{Shujuan Chen\corref{cor1}}
\ead{sc2331@cam.ac.uk}
\author[1]{Ying Jin}

\cortext[cor1]{Corresponding author.}

\affiliation[1]{organization={Martin Centre for Architectural and Urban Studies, University of Cambridge},
            addressline={1--5 Scroope Terrace},
            city={Cambridge},
            postcode={CB2 1PX},
            country={United Kingdom}}

\begin{abstract}
Short-term traffic forecasting supports real-time monitoring and control of road networks, and graph attention networks (GAT) are the standard means of representing spatial dependence in these models. GAT layers are widely described as capturing the influence of neighbouring locations, but this is seldom verified, because the attention weights offered in support cannot be compared against any measured quantity. That leaves two questions open, how the model should be explained and which of its components are necessary. We address this by adding a gate to the GAT layer which learns, at every sensor and every time step, what share of a sensor's updated state is drawn from its neighbours rather than from itself. Regularising the gate withdraws neighbour information progressively and thereby provides a graded form of ablation. We apply the gated GAT to ST-MetaNet, whose encoder and decoder each place one GAT layer between two recurrent layers, and train it on one calendar year of records from 498 loop detectors on the strategic road network of England. The gate assigns a larger share of neighbour information to sensors carrying heavier traffic and follows the daily and weekly cycle of travel, consistent with adjacent locations being more strongly coupled when busy. Mild regularisation improves accuracy slightly, and accuracy declines at higher strengths as the penalty withdraws information the model needs. The encoder gate closes before the decoder gate, but direct ablation qualifies that ordering. Removing either GAT layer alone leaves accuracy at least as good as keeping both, whereas removing both degrades it substantially, so the two layers are largely redundant rather than either being indispensable. The gated GAT therefore yields a modest accuracy gain, an explanation of where and when spatial information flows, and evidence on which layers the architecture requires.
\end{abstract}

\begin{keyword}
traffic forecasting \sep graph neural networks \sep model explainability \sep ablation \sep strategic road network
\end{keyword}

\end{frontmatter}

\section{Introduction}
\label{sec:intro}

Short-term traffic forecasting supports ramp metering, incident detection, variable message signing and the provision of journey time information on national road networks \citep{Introimportanceoftrafficprediction}. Because these forecasts are generated continuously and acted upon within minutes, the models that produce them are expected to satisfy requirements beyond predictive accuracy alone. Two are particularly relevant here. The first is explainability, since a model whose internal reasoning cannot be examined cannot be diagnosed when it fails, and confidence established on aggregate accuracy does not necessarily extend to the conditions under which the model was never evaluated \citep{blackbox, explainability, StopExplaining}. The second is structural economy, since a forecaster operating continuously across a national sensor network incurs computational cost at every interval, and each additional layer increases the latency between measurement and decision, so that components which are retained should be demonstrably necessary \citep{PruningConnections, LotteryTicket}. Both requirements converge on a single question. These models are constructed to exchange information between locations, on the assumption that conditions at one point of a road network depend on conditions at the points adjacent to it. Whether this exchange contributes materially, and where and when it does so, determines both how the model should be explained and how much of its structure is required.

Deep learning now dominates this task, and its gains have come from a succession of architectural components developed over the past decade \citep{GNN4TFreview, TFflowReview}. Recurrent neural networks (RNN) capture the temporal structure of a sensor series \citep{RNN, LSTM, GRU}, and graph neural networks (GNN) extend the treatment to the spatial structure of the road network \citep{GNN, GCN}. Current models combine these blocks into spatio-temporal architectures that represent both kinds of structure at once. They include diffusion convolutional recurrent networks, spatio-temporal graph convolutional networks and their attention-based successors \citep{DCRNN, STGCN, STGAT, AGCRN}. Graph attention networks (GAT) weight each neighbour by a learned coefficient rather than by a fixed adjacency, which allows the strength of a spatial dependence to vary with the data \citep{GAT, Attention}. The GAT layer has consequently become the standard way of representing spatial dependence, placed between recurrent layers that handle the temporal dependence. ST-MetaNet is a representative model with this sandwich structure \citep{Pan2019, Pan2022}.

Alongside these gains, a body of work has developed methods for explaining what a graph network has learned. GNNExplainer and its parameterised successor identify the subgraph and node features that most influence a prediction \citep{GNNExplainer, PGExplainer}. GraphMask learns a binary mask over edges and reports which connections can be deleted without changing the output \citep{GraphMask}. A taxonomic survey groups these approaches by whether they perturb inputs, propagate gradients, or fit a surrogate \citep{GNNExplainSurvey}. In transport specifically, García-Sigüenza et al.~\citep{GarciaSiguenza2023} apply several of these techniques to an adaptive graph convolutional recurrent network and report that the model draws on simpler information at low flows and richer information as flow rises. Tygesen et al.~\citep{Unboxing} use neural relational inference to recover a latent interaction graph and find that information is routed towards congested locations. A separate line of work in natural language processing has examined whether attention weights can be read as explanations at all, and has produced both critical and defending accounts \citep{AttentionNotExplanation, IsAttentionInterpretable, AttentionNotNotExplanation, SaliencyElephant}.

Taken together, these approaches leave both requirements unmet for the spatial component of traffic forecasting models. Explainability is unmet because the explanations offered cannot be validated. An explanation method returns a quantity for each edge or node pair which is then read as importance, yet in most settings no external measurement of the phenomenon it claims to describe is available, so the interpretation rests on whether the resulting pattern appears plausible. Traffic is unusual in this respect, because the physical state of the network is observed at every sensor and every interval and therefore supplies a reference against which an internal quantity can be compared, but this comparison is rarely performed. Structural economy is unmet because a quantity read from a trained model is not evidence that the component producing it is required. Studies in image classification have shown that measures of this kind can disagree with the effect of actually removing the component they score \citep{SingleDirections, ROAR, SanityChecks}, and the pruning literature has established that trained networks routinely contain components which can be deleted without loss because the remainder compensates \citep{PruningConnections, LotteryTicket}. Whether an internal reliance measure in a traffic forecasting model behaves in the same way has not been established, and until it has, such a measure cannot be used to determine which parts of the model are needed.

In this study we address both requirements within a single setting. We add a gate to the GAT layer of ST-MetaNet, a representative spatio-temporal forecaster whose encoder and decoder each place one GAT layer between two recurrent layers. The gate learns, at every sensor and every time step, what share of a sensor's updated state is drawn from its neighbours rather than from itself, and regularising the gate withdraws that neighbour information progressively. This design leads to three contributions. First, because the gate is part of the forecaster rather than a probe applied to it afterwards, it is trained with the model and its effect on forecast accuracy can be observed directly, which we do across the full range of the regularisation strength and which yields a modest improvement under mild regularisation. Second, the gate provides a continuous and spatially resolved account of where and when spatial information flows within the model, which we validate against the traffic state observed independently at the same sensors, a reference available in traffic forecasting and in few other settings. Third, regularising the gate provides a graded form of ablation whose implications can be checked against the direct alternative of deleting each GAT layer and retraining, which establishes where the internal account agrees with removal and where it does not, and shows the two GAT layers to be largely redundant with one another. To our knowledge this is the first study to evaluate an internal measure of spatial information flow in a traffic forecasting model both against the measured traffic state and against direct ablation of the same layers.

\section{Methodology}
\label{sec:method}

\begin{figure}[htbp]
    \centering
    \makebox[\textwidth][c]{\includegraphics[width=1.08\textwidth]{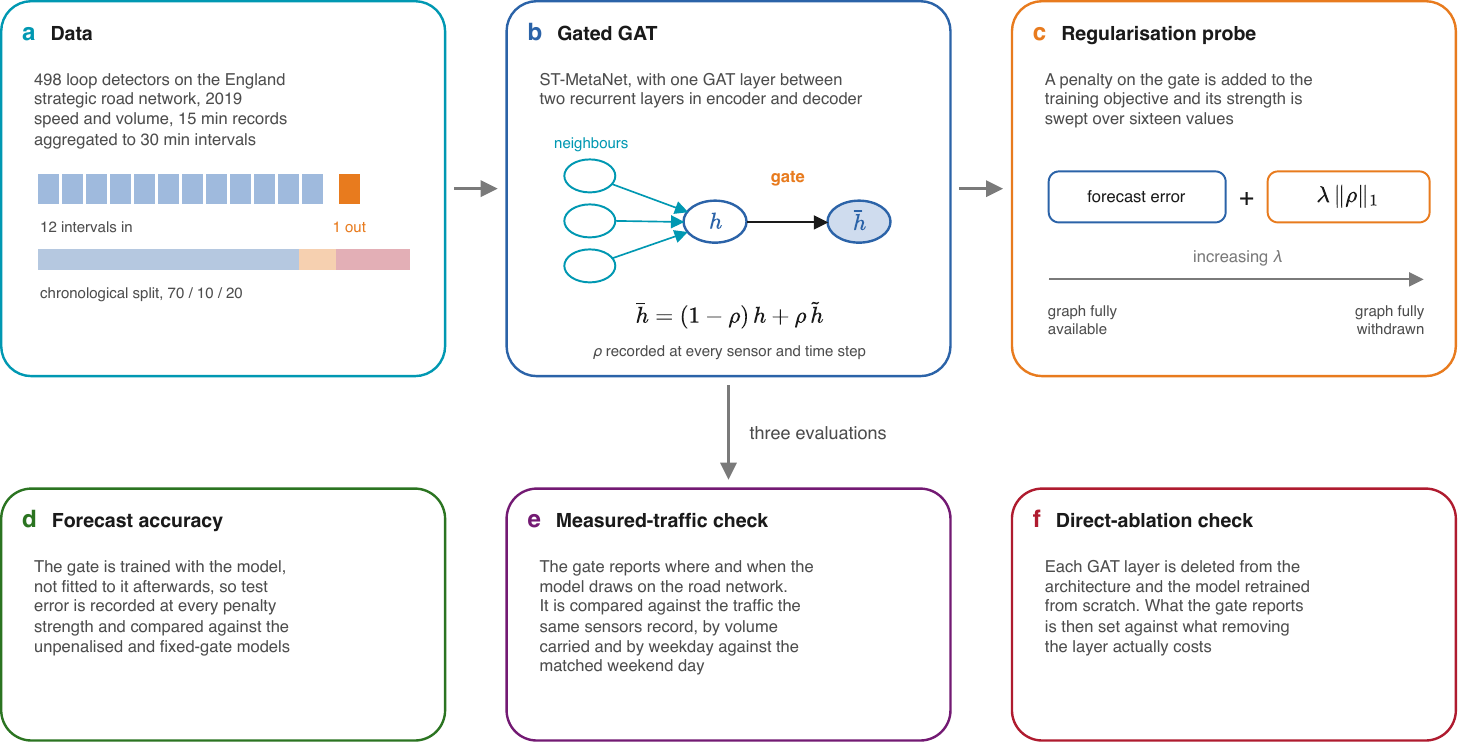}}
    \caption{\textbf{Study design.} \textbf{a}, The data, one calendar year of speed and volume from 498 loop detectors on the England strategic road network, aggregated to 30 minute intervals and divided chronologically. \textbf{b}, The model, ST-MetaNet with a gate inserted at the node update of each graph attention layer, so that the share of the update taken from neighbouring sensors is recorded at every sensor and every time step. \textbf{c}, The probe, a penalty on the gate added to the training objective and swept over sixteen strengths, which withdraws neighbour information progressively and so provides a graded ablation. \textbf{d} to \textbf{f}, The three evaluations this design supports. Results are reported in Figures~\ref{fig:accuracy} to~\ref{fig:ablation}.}
    \label{fig:concept}
\end{figure}

The framework comprises three components, summarised in Figure~\ref{fig:concept} and presented below in the order in which they were developed.

The first is the model. We take a published spatio-temporal forecaster, ST-MetaNet, and modify a single element of it: the fixed weight that combines each sensor's own state with the aggregate of its neighbours is replaced by a gate computed from the two states being combined, so that the share of the update drawn from the road network is recorded for every sensor at every timestep. Adding a penalty on the gate to the training objective serves a second purpose, since the model can then be observed under a controlled and progressive reduction in the neighbour information available to it.

The second is the data. The model is trained and evaluated on one calendar year of loop detector records from 498 sensors on the strategic road network of England, aggregated to 30 minute intervals, predicting volume and speed one interval ahead from twelve intervals of history.

The third is the experimental design, which is what makes the gate a measurement rather than an illustration. Because the gate is a quantity the model reports about itself, it establishes nothing in isolation, and we therefore evaluate it against two references that exist outside the model. The first is the measured traffic state, recorded independently at every sensor and every interval. The second is direct ablation, in which each graph layer is deleted from the architecture and the model retrained from scratch, so that the consequence of removing a layer is observed rather than inferred.

\subsection{Forecasting model}
\label{subsec:model}

\subsubsection{Base model}

\begin{figure}[htbp]
    \centering
    \includegraphics[width=\textwidth]{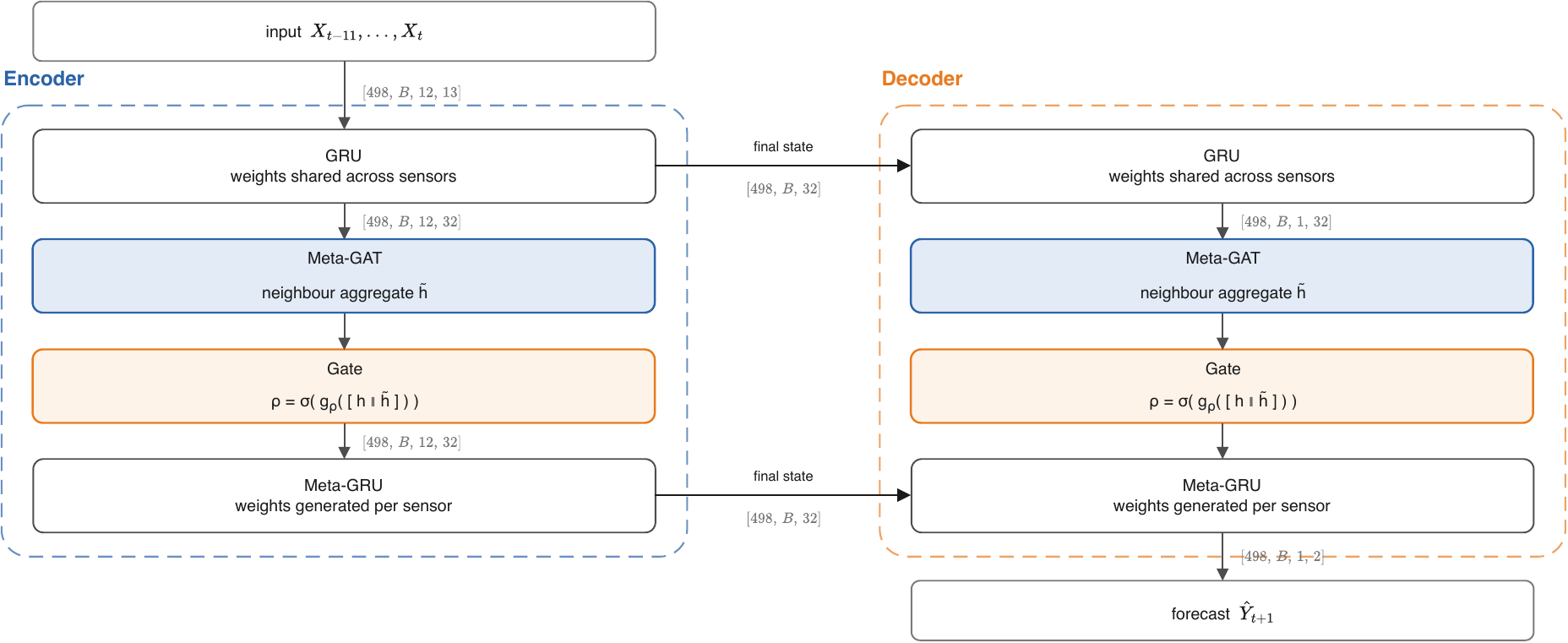}
    \caption{\textbf{Model architecture and the position of the gate.} Encoder and decoder each hold one meta graph attention layer, and the gate sits at the node update of each. Dashed outlines mark the two gates, which are the only quantities the penalty acts on. The penalty applies equal weight to every gate value, and the encoder gate is evaluated at twelve intervals against the decoder gate's one, so twelve of every thirteen penalised values belong to the encoder. Meta-knowledge learners generate the parameters of the graph attention and of the second recurrent layer from node and edge geography. The dual graph carries the 498 sensors as nodes with separate incoming and outgoing edge sets, whose outputs are averaged. Bracketed quantities give tensor shapes, with $B$ the batch size.}
    \label{fig:arch}
\end{figure}

The forecaster is ST-MetaNet \citep{Pan2019}, a sequence to sequence model in which an encoder compresses the twelve input intervals into a latent state and a decoder produces the forecast. Encoder and decoder have the same three layers, applied in the same order (Figure~\ref{fig:arch}). A gated recurrent unit with weights shared across all sensors encodes the temporal signal \citep{GRU}. A meta graph attention layer then exchanges information between sensors. A second recurrent layer, whose weights are generated per sensor from that sensor's geographic attributes, produces the layer output.

The meta graph attention layer differs from a standard graph attention layer in how the attention coefficients are produced \citep{GAT}. Rather than learning one set of attention parameters for the whole graph, a fully connected network maps the geographic attributes of a node pair to the weights and biases used to score that pair. The score is normalised across the neighbours of each node with a softmax, and the neighbour states are combined using the normalised scores. The incoming and outgoing dual graphs are processed separately and their outputs are averaged. All hidden states have 32 dimensions. Departures from the original ST-MetaNet implementation are listed in Appendix~S7.

\subsubsection{Gated GAT}

In the original formulation the updated state of a node is a fixed combination of the node's own state and the aggregate of its neighbours, controlled by a single learned scalar for each graph. That scalar is constant across sensors and across time, so it cannot record that a given sensor draws more on its neighbours during a peak than at night. We replace it with a gate whose value is produced from the two states being combined.

Let $h_t^{(i)}$ denote the state of sensor $i$ at time $t$ entering the graph layer, and let
\begin{equation}
\tilde{h}_t^{(i)} = \mathrm{ReLU}\Big( \sum_{j \in \mathcal{N}(i)} \alpha^{(ij)} \odot h_t^{(j)} \Big)
\end{equation}
denote the aggregate of its neighbours. The attention coefficients $\alpha^{(ij)} \in \mathbb{R}^{32}$ are normalised across neighbours separately in each hidden dimension, and $\odot$ denotes the elementwise product. The gate is
\begin{equation}
\rho_t^{(i)} = \sigma\Big( g_\rho\big( [\, \tilde{h}_t^{(i)} \,\|\, h_t^{(i)} \,] \big) \Big),
\end{equation}
where $\|$ denotes concatenation, $g_\rho$ is a fully connected network with hidden widths 16 and 2, and $\sigma$ is the logistic function. The updated state is
\begin{equation}
\bar{h}_t^{(i)} = \big( 1 - \rho_t^{(i)} \big)\, h_t^{(i)} + \rho_t^{(i)}\, \tilde{h}_t^{(i)} .
\end{equation}
The gate therefore takes a value in the open unit interval for every sensor, every timestep and each of the two dual graphs. A value near zero means the sensor's updated state ignores its neighbours, and a value near one means the sensor's own state is discarded in favour of them. The graph attention layers of ST-MetaNet are thereby gated, and we refer to the resulting model as Gated ST-MetaNet.

\subsubsection{Regularisation}

Adding a penalty on the gate to the training objective allows the model to be observed under a controlled reduction in how much it may use the graph. The objective is
\begin{equation}
\mathcal{L} = \mathcal{L}_{\mathrm{train}} + \lambda \, \frac{1}{|\mathcal{R}|} \sum_{r \in \mathcal{R}} \big| \rho_r \big| ,
\end{equation}
where $\mathcal{L}_{\mathrm{train}}$ is the mean squared error on standardised observed values and $\mathcal{R}$ indexes every gate value produced in a batch. The index set covers all 498 sensors, all twelve encoder intervals and the single decoder interval, and both dual graphs. Every gate value therefore carries the same weight in the penalty. The two layers do not receive the same total pressure, because the encoder gate is evaluated at twelve intervals and the decoder gate at one, so twelve of every thirteen penalised values belong to the encoder. This asymmetry is a property of the penalty as specified and is carried forward into the interpretation of the results. At $\lambda = 0$ the objective is the unmodified forecasting loss. As $\lambda$ grows the model is pushed towards ignoring the graph entirely, which reduces it to two stacked recurrent layers.

The value of the probe is the path traced between these two limits. Under a shrinkage penalty of this form, the quantities an objective depends on most tend to persist longest, which is the basis on which coefficient paths are read in penalised regression \citep{Lasso, Ridge}. We evaluate sixteen values of $\lambda$ from $0$ to $10^{4}$.

\subsection{Study network and data}
\label{subsec:data}

\subsubsection{Highway network}

The study network is the England trunk road graph that we introduced for an earlier study of interurban travel demand \citep{mukara1}, reused here without modification. It holds 181 nodes and 498 directed edges. A node is a junction on the National Highways Strategic Road Network \citep{national_highways}, and most junctions lie close to a major town or city. An edge is one direction of one of the 249 trunk road segments joining those junctions, and each edge carries a single loop detector from the National Highways Traffic Information System (TRIS) \citep{TRIS2024}. The construction rules are set out in full in that earlier study. In summary, junctions and segments were selected by hand so that the graph follows the trunk road corridors between towns and cities rather than bypasses and grade separated exits, and where a segment offered more than one detector, the detector nearest the midpoint of the segment with the lower rate of missing records was taken. The resulting network is shown in Figure~\ref{fig:data}a.

Because the traffic measurements sit on road segments rather than on junctions, the model operates on the dual of this graph. Each of the 498 segments becomes a node, and two segments are joined when they meet at a junction. Direction is preserved by separating the connections into an incoming graph, in which segments are joined through the junction at their upstream end, and an outgoing graph, in which they are joined through the junction at their downstream end. Each graph contains 2{,}646 directed edges. A junction carries between two and ten segments, with a mean of 5.50 (Figure~\ref{fig:data}f). Each dual-graph node therefore has between one and nine neighbours in each direction, with a mean of 5.31. Node attributes in the dual graph are the sensor coordinates, the length and free-flow driving duration of the segment, the coordinates of the two junctions it connects, and the number of segments meeting at each of those junctions, giving ten attributes. Edge attributes in the dual graph are the coordinates of the shared junction. All attributes are standardised before use.

\subsubsection{Traffic measurements}

Speed and volume are recorded at 15 minute resolution by inductive loop detectors and published through TRIS, the traffic information service operated by National Highways \citep{NationalHighways2024}. The service has run since 2014 and now covers 19{,}364 detectors across England. Records for the 498 detectors of the study network were retrieved through the TRIS application programming interface \citep{TRIS2024}, as in our earlier study of the same network \citep{mukara1}.

\begin{figure}[htbp]
    \centering
    \includegraphics[width=\textwidth]{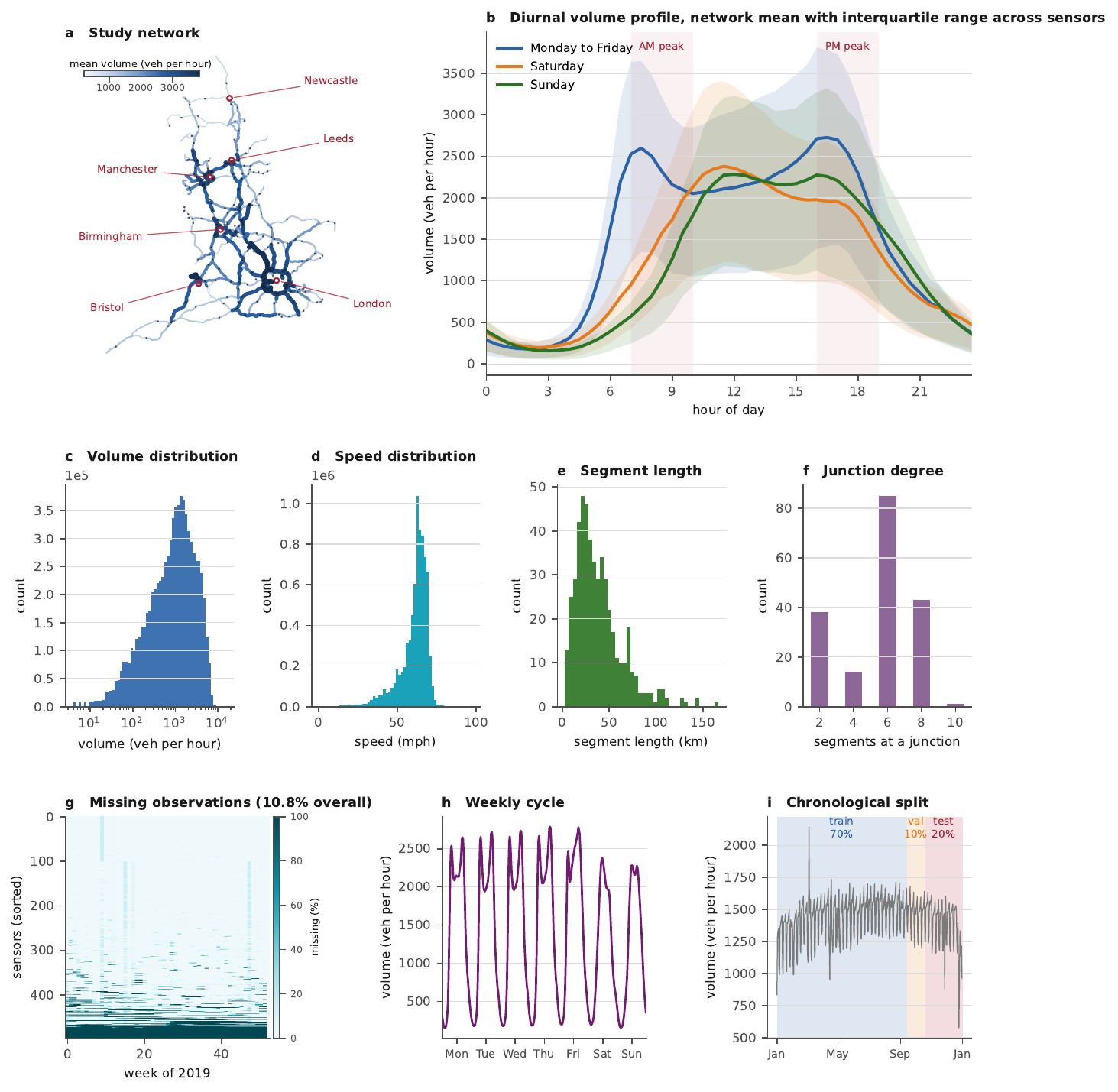}
    \caption{\textbf{Study network and traffic measurements, 2019.}
\textbf{a}, The 498 sensored trunk road segments, shaded and weighted by mean flow, with named cities marked for orientation.
\textbf{b}, Diurnal flow profile by day type, showing the network mean with the interquartile range across the 498 sensors. Shaded bands mark the morning and evening peak windows.
\textbf{c}, Distribution of 30 minute flow across all sensors and intervals.
\textbf{d}, Distribution of time mean speed.
\textbf{e}, Distribution of segment length.
\textbf{f}, Number of segments meeting at each of the 181 junctions.
\textbf{g}, Missing observations by sensor and week, sensors sorted by overall completeness. The overall rate is 10.8\% at the 30 minute resolution.
\textbf{h}, Weekly cycle of network mean flow.
\textbf{i}, Chronological division into training, validation and test periods.}
    \label{fig:data}
\end{figure}

We use the calendar year 2019 for all 498 sensors, which gives 35{,}040 records of speed and volume at each sensor. The year 2019 is the most recent year of ordinary operation before the travel restrictions associated with the COVID-19 pandemic. All days are retained, including weekends and public holidays, because the contrast between a weekday and a weekend day is one of the quantities the analysis tests. Across the full year, 10.53\% of volume records and 10.73\% of speed records are missing. Missingness is concentrated in a small group of sensors rather than spread evenly, and a few sensors are unavailable for whole weeks (Figure~\ref{fig:data}g). A further 0.32\% of records report a volume of zero, which arises when no vehicle is detected during the interval. Counting those as missing as well gives the 10.8\% rate shown in Figure~\ref{fig:data}g. Mean flow across the network is 1{,}426 vehicles per hour and mean speed is 60.6 miles per hour.

Fifteen minute counts on a single loop detector are noisy, and a large part of the variation between consecutive intervals reflects the arrival process rather than any change in the state of the network. We therefore aggregate to 30 minute intervals, which reduces this variation while retaining the within-peak structure the analysis depends on. Volume is expressed as a flow rate in vehicles per hour and speed as a time mean in miles per hour, so that results at different aggregation intervals are directly comparable. Aggregation gives 17{,}520 intervals in the year.

The series is divided along the time axis into 70\% training, 10\% validation and 20\% test, without shuffling. Training covers 1 January to 13 September 2019, validation covers 13 September to 19 October, and the test period covers 20 October to 31 December, which is 73 days (Figure~\ref{fig:data}i). Splitting by time rather than at random prevents an interval adjacent to a test interval from entering training. Speed and volume are standardised using the mean and standard deviation of the training period only.

A missing measurement is recorded in a binary indicator channel and its value is replaced by a constant. Every loss and every reported error metric is computed only over intervals with an observed value, using this indicator. Input windows containing missing measurements are retained, so the model is trained under the same pattern of gaps it would meet in operation. Alongside the two traffic channels and their two indicator channels, each input carries the time of day scaled to the unit interval, a seven dimensional indicator of the day of the week, and an indicator of English public holidays, giving thirteen input channels.

Samples are formed with a sliding window of twelve input intervals and one output interval, at a stride of one interval. This gives 12{,}252 training, 1{,}740 validation and 3{,}492 test samples.

\subsection{Experimental design}
\label{subsec:design}

The gate is a quantity the model produces about itself, so on its own it establishes nothing. We test it in two independent ways.

The first test asks whether the gate is associated with the measured traffic state. If the gate records reliance on neighbouring sensors, then it should be higher where traffic is heavier, because a busy location on a trunk road is more strongly coupled to the locations upstream and downstream of it. It should also differ between a weekday and a matched weekend day at the times when weekday travel differs most, namely the morning and evening peaks, and not at times when the two days are similar. Neither quantity is supplied to the gate as an input feature, and neither is supervised. The association is nonetheless not fully independent of the model, because the gate is computed from the node's own hidden state, which encodes that sensor's recent volume history. A gate that varied with volume for that reason alone would produce the same correlation. What the comparison establishes is that the learned gate is ordered by traffic load rather than being flat or arbitrary, not that the ordering is caused by reliance on neighbours. We take the mean gate value across the twelve encoder intervals for each sensor and correlate it with the flow observed at that sensor at the prediction time, using the Spearman rank correlation. We repeat the correlation against the distance from each sensor to central London as a comparison quantity that should not carry the association if the gate is tracking traffic rather than geography. Four prediction times are used, at 03:44, 09:44, 15:44 and 21:44, on Monday 4 November and Sunday 10 November 2019. Sensors recording fewer than 40 vehicles per hour are excluded from the correlation, because speed and flow are both unreliable at those counts.

The second test asks whether the order in which the two graph layers give up reliance under the penalty identifies which layer the architecture needs. We train four architectures at each of four aggregation intervals: the full model, the model with the encoder graph layer removed, the model with the decoder graph layer removed, and a model with no graph layer, which reduces to two stacked recurrent layers. Each variant is trained from scratch under the same protocol. We repeat the whole set with a clipped linear gate, $\rho = \min(\max(x,0),1)$, in place of the logistic gate, to establish whether the outcome depends on that choice. Appendix~S9 documents which form of the clipped gate each reported run used. Unlike the logistic gate, the clipped gate can take the value zero exactly.

\subsection{Training and evaluation protocol}
\label{subsec:training}

Training minimises the objective above with the Adam optimiser at a learning rate of $0.01$, a batch size of 32 samples, and gradients clipped elementwise at $5$. The number of passes over the training set is set separately for each aggregation interval so that the number of gradient updates is held between 766 and 768, which gives one pass at the 15 minute interval, two at 30 minutes, four at 60 minutes and twelve at 180 minutes. Every configuration is trained once from a fixed random seed. Appendix~S4 lists every configuration reported here with its aggregation interval, gate nonlinearity, penalty strength and architecture. The pipeline is deterministic, and two separately executed runs of the same configuration returned identical results to sixteen significant figures, so repeated execution contributes no variation. What is not quantified is the variation that a different initialisation would produce.

Forecast accuracy is reported as the mean absolute error and the root mean squared error, computed over observed values only, separately for volume and for speed. The training code records volume as the mean of the constituent 15 minute counts and speed as their sum, so volume is multiplied by four to give a flow rate in vehicles per hour and speed is divided by the number of 15 minute periods in the aggregation interval to give miles per hour. Both conversions are exact on intervals with no missing constituent measurement. The reported metric is the mean of per-batch means. Because the final batch of each split is smaller than the others, this places slightly more weight on the last few samples than a pooled mean would. Recomputing as a mean pooled over all observations raises the volume error by between 0.31 and 0.55 vehicles per hour, in the same direction for every configuration.

To assess whether a difference in accuracy between two configurations is larger than the variation expected from the composition of the test period, we compare the saved test predictions of the two models directly. This comparison uses errors pooled over all observations rather than the batch-mean metric reported in the tables, so its absolute values differ from them by the amount given above. Absolute errors are resampled at the level of whole days, using 10{,}000 paired bootstrap replicates over the 73 test days. Resampling whole days rather than individual intervals respects the strong autocorrelation within a day. This procedure describes variation arising from which days fall in the test period. It does not describe variation between training runs.

All experiments were run on a single workstation using Python 3.9.18, TensorFlow 2.10.1 and DGL 1.1.2. The archived training scripts disable GPU visibility, so all training times reported below are processor times on that workstation. Reporting follows the TRIPOD-AI recommendations for prediction model studies as far as they apply to a study without human participants.

\section{Results}
\label{sec:results}

\subsection{Forecast accuracy under the penalty}

\begin{figure}[htbp]
    \centering
    \includegraphics[width=\textwidth]{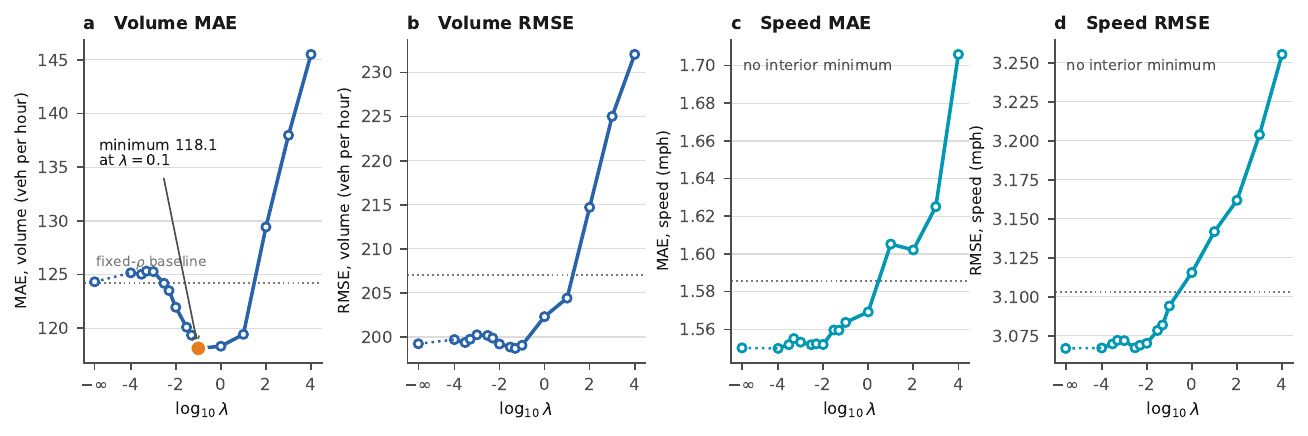}
    \caption{\textbf{Forecast accuracy against penalty strength at the 30 minute interval.} Each panel reports one metric on the held-out test period. The dotted line marks the fixed-gate ST-MetaNet baseline for that metric. The leftmost point is the unpenalised model, plotted at a detached position because $\log_{10} 0$ is undefined, and joined by a dotted segment. \textbf{a}, Mean absolute error for volume, with the minimum marked. \textbf{b}, Root mean squared error for volume. \textbf{c}, Mean absolute error for speed. \textbf{d}, Root mean squared error for speed.}
    \label{fig:accuracy}
\end{figure}

We first evaluated forecast accuracy across sixteen penalty strengths at the 30 minute interval, against the fixed-gate ST-MetaNet baseline (Figure~\ref{fig:accuracy}, Table~\ref{tab:accuracy}). Volume error falls and then rises as the penalty strengthens. Mean absolute error for volume is 124.3 vehicles per hour at $\lambda = 0$, reaches a minimum of 118.1 at $\lambda = 0.1$, and rises to 145.5 at $\lambda = 10^{4}$. The baseline attains 124.2. Volume root mean squared error is flat across the lower half of the range, varying between 198.7 and 200.2 for every penalty strength up to $\lambda = 0.1$, and rises thereafter.

\begin{table}[!ht]
\centering
\small
\caption{\textbf{Forecast accuracy against penalty strength at the 30 minute interval.} All values are on the held-out test period of 73 days. Volume errors are in vehicles per hour and speed errors in miles per hour. The fixed-gate ST-MetaNet baseline is given for reference in the first row. The lowest value in each column among the penalised models is shown with the penalty strength that produced it in the final row.}
\label{tab:accuracy}
\begin{tabular}{lrrrrr}
\toprule
Model & $\log_{10}\lambda$ & MAE volume & RMSE volume & MAE speed & RMSE speed \\
\midrule
ST-MetaNet, fixed gate & -- & 124.19 & 207.00 & 1.586 & 3.103 \\
\midrule
\multicolumn{6}{l}{\textit{Gated ST-MetaNet, logistic gate}} \\
\hspace{1em}no penalty & $-\infty$ & 124.33 & 199.22 & 1.550 & 3.067 \\
\hspace{1em}$\lambda = 10^{-4}$ & $-4.00$ & 125.16 & 199.70 & 1.550 & 3.067 \\
\hspace{1em}$\lambda = 3\times10^{-4}$ & $-3.52$ & 125.03 & 199.36 & 1.552 & 3.070 \\
\hspace{1em}$\lambda = 5\times10^{-4}$ & $-3.30$ & 125.32 & 199.74 & 1.555 & 3.072 \\
\hspace{1em}$\lambda = 10^{-3}$ & $-3.00$ & 125.28 & 200.24 & 1.553 & 3.072 \\
\hspace{1em}$\lambda = 3\times10^{-3}$ & $-2.52$ & 124.19 & 200.16 & 1.552 & 3.067 \\
\hspace{1em}$\lambda = 5\times10^{-3}$ & $-2.30$ & 123.51 & 199.87 & 1.552 & 3.069 \\
\hspace{1em}$\lambda = 10^{-2}$ & $-2.00$ & 121.96 & 199.19 & 1.552 & 3.070 \\
\hspace{1em}$\lambda = 3\times10^{-2}$ & $-1.52$ & 120.10 & 198.85 & 1.560 & 3.078 \\
\hspace{1em}$\lambda = 5\times10^{-2}$ & $-1.30$ & 119.36 & \textbf{198.70} & 1.560 & 3.082 \\
\hspace{1em}$\lambda = 10^{-1}$ & $-1.00$ & \textbf{118.13} & 199.05 & 1.564 & 3.094 \\
\hspace{1em}$\lambda = 1$ & $0.00$ & 118.33 & 202.30 & 1.569 & 3.116 \\
\hspace{1em}$\lambda = 10$ & $1.00$ & 119.44 & 204.40 & 1.605 & 3.142 \\
\hspace{1em}$\lambda = 10^{2}$ & $2.00$ & 129.43 & 214.71 & 1.602 & 3.162 \\
\hspace{1em}$\lambda = 10^{3}$ & $3.00$ & 137.97 & 225.03 & 1.625 & 3.204 \\
\hspace{1em}$\lambda = 10^{4}$ & $4.00$ & 145.50 & 232.07 & 1.706 & 3.255 \\
\bottomrule
\end{tabular}
\end{table}

Speed error behaves differently. Mean absolute error for speed rises essentially without interruption, from 1.550 miles per hour at $\lambda = 0$ to 1.706 at $\lambda = 10^{4}$, with no minimum distinguishable at the reported precision. The pattern of a minimum away from zero therefore holds for volume and not for speed. Validation volume error reaches its minimum at the same penalty strength as test volume error, at $\lambda = 0.1$.

The paired day-block bootstrap places the volume difference outside the range attributable to the composition of the test period. Against $\lambda = 0$, the model at $\lambda = 0.1$ has a volume error lower by 5.96 vehicles per hour, with a 95\% interval of 5.15 to 6.89 lower. Its speed error is higher by 0.013 miles per hour, with a 95\% interval of 0.008 to 0.018 higher. At the milder setting of $\lambda = 0.01$ the volume error is lower by 2.23 vehicles per hour, with a 95\% interval of 1.74 to 2.80 lower, and the speed difference of 0.001 miles per hour has an interval from 0.002 lower to 0.004 higher, which includes zero. The volume gain at $\lambda = 0.1$ is about one third of the standard deviation of volume error across the 73 test days, which is 17.4 vehicles per hour around a mean of 124.5. Full bootstrap distributions are given in Supplementary Figure~S2.

\subsection{Association between the gate and the measured traffic state}

\begin{figure}[htbp]
    \centering
    \includegraphics[width=\textwidth]{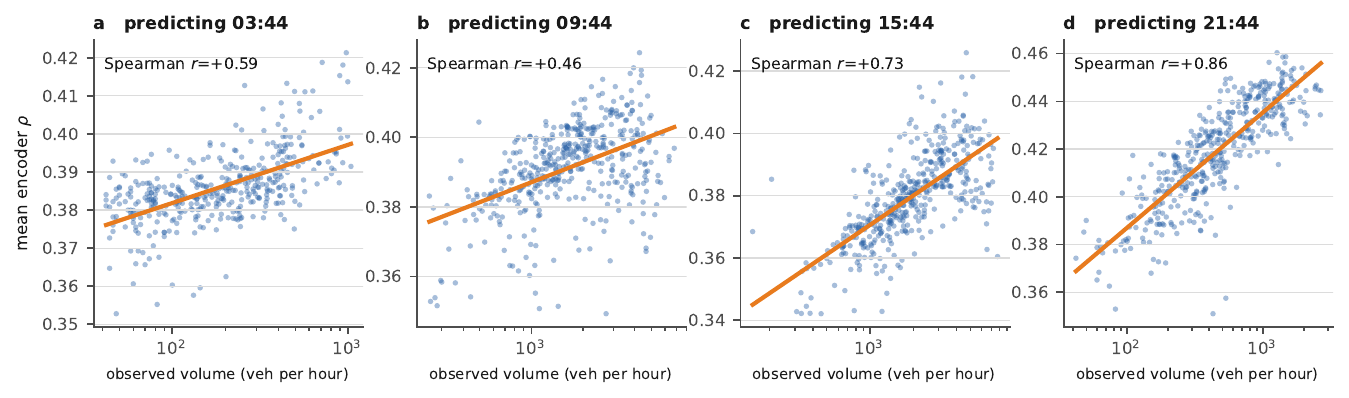}
    \caption{\textbf{The gate against measured flow.} Each point is one sensor, on Monday 4 November 2019, under the unpenalised model, at the four prediction times \textbf{a} to \textbf{d}. The vertical axis is the mean gate value across the twelve encoder intervals and the horizontal axis is the flow observed at that sensor at the prediction time, on a logarithmic scale, with a fitted line. Between 402 and 453 sensors are plotted depending on the panel. Sensors recording fewer than 40 vehicles per hour are excluded. Correlations are Spearman rank correlations. The corresponding comparison against distance from central London is given in Appendix~S6.}
    \label{fig:load}
\end{figure}

We next examined whether the gate is associated with quantities measured independently of the model. Reliance on neighbouring sensors rises with the flow carried at a sensor at every prediction time examined (Figure~\ref{fig:load}). On Monday 4 November 2019 the Spearman correlation between the mean encoder gate and observed flow is 0.86 at 21:44, 0.73 at 15:44, 0.59 at 03:44 and 0.46 at 09:44, with $P < .001$ in every case and between 402 and 453 sensors contributing. On the matched Sunday the corresponding correlations are 0.85, 0.58, 0.71 and 0.52.

The comparison quantity carries no association of comparable size. Correlations between the same gate values and the distance from each sensor to central London reach at most 0.19 in absolute value across the eight combinations of day and time, against 0.46 as the smallest correlation with flow. The full set is given in Appendix~S6. The sign changes four times. Three of the eight are not distinguishable from zero, with $P$ values of .56, .16 and .10, and the remaining five reach $P < .05$ despite their small magnitude.

The gate also separates a weekday from the matched weekend day, and does so only at the peaks (Table~\ref{tab:daycontrast}). Paired across the 498 sensors, the mean encoder gate on Monday exceeds the matched Sunday value by 0.0143 when predicting 09:44 and by 0.0112 when predicting 21:44, with 95\% intervals of 0.0134 to 0.0151 and 0.0102 to 0.0121 and $P < .001$ in both cases. At 03:44 the difference is 0.0010, with a 95\% interval of 0.0006 to 0.0014 and $P < .001$, about one fourteenth of the difference at the morning peak. At 15:44 it is 0.0003, with a 95\% interval from $-0.0004$ to $0.0010$ and $P = .24$.

\begin{table}[!ht]
\centering
\small
\caption{\textbf{Mean encoder gate on a weekday against the matched weekend day.} Values are means across the 498 sensors, averaged over the twelve encoder intervals, under the unpenalised model. Monday is 4 November 2019 and Sunday is 10 November 2019. Differences are paired across sensors. Intervals are 95\% bootstrap intervals from 10{,}000 resamples and $P$ values are from the Wilcoxon signed-rank test.}
\label{tab:daycontrast}
\begin{tabular}{lrrrrr}
\toprule
Prediction time & Monday & Sunday & Difference & 95\% interval & $P$ \\
\midrule
03:44 & 0.3847 & 0.3837 & $+0.0010$ & $+0.0006$ to $+0.0014$ & $<.001$ \\
09:44 & 0.3904 & 0.3761 & $+0.0143$ & $+0.0134$ to $+0.0151$ & $<.001$ \\
15:44 & 0.3780 & 0.3777 & $+0.0003$ & $-0.0004$ to $+0.0010$ & .24 \\
21:44 & 0.4171 & 0.4059 & $+0.0112$ & $+0.0102$ to $+0.0121$ & $<.001$ \\
\bottomrule
\end{tabular}
\end{table}

\begin{figure}[htbp]
    \centering
    \includegraphics[width=\textwidth]{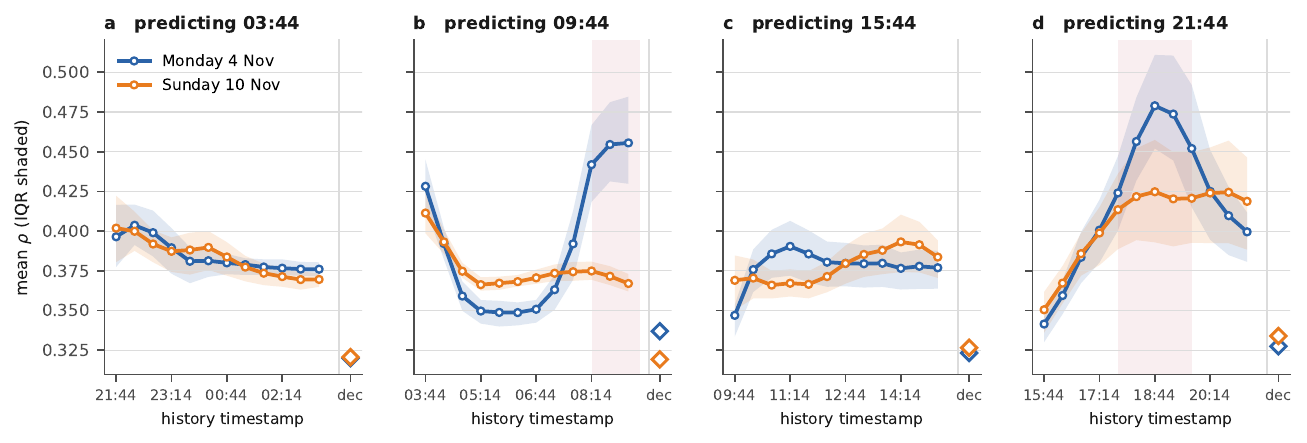}
    \caption{\textbf{Gate values across the history window, weekday against matched weekend day.} Each panel corresponds to one prediction time under the unpenalised model. Lines show the mean across the 498 sensors and shaded bands the interquartile range. The horizontal axis gives the clock time of each of the twelve encoder intervals. The diamond to the right of the vertical rule is the single decoder interval. Shaded regions in \textbf{b} and \textbf{d} mark the intervals covering the morning and evening peaks.}
    \label{fig:lag}
\end{figure}

\begin{figure}[htbp]
    \centering
    \includegraphics[width=\textwidth]{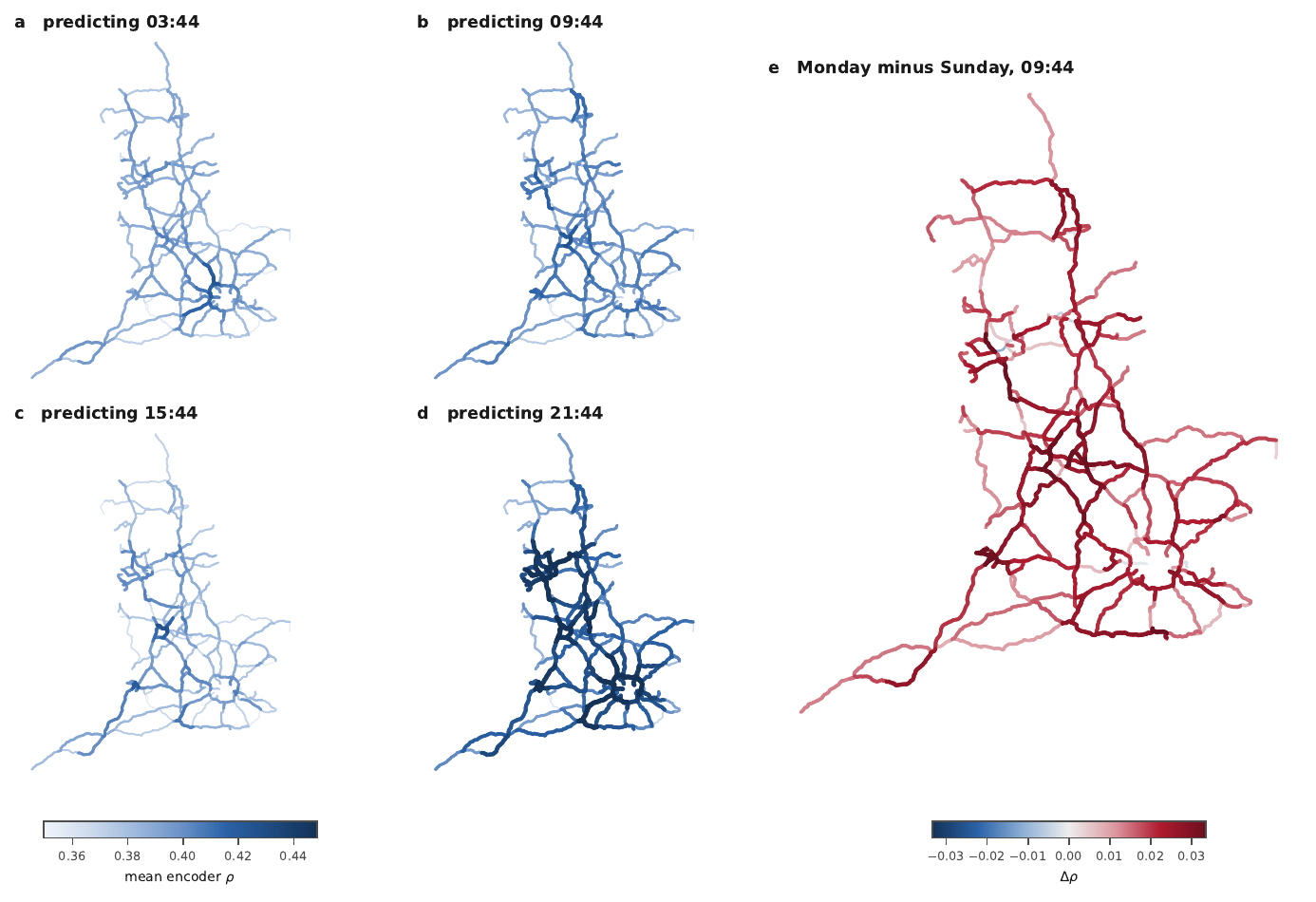}
    \caption{\textbf{Gate values mapped onto the network.} Mean encoder gate value for each of the 498 segments under the unpenalised model, with line width and colour both carrying the value. \textbf{a} to \textbf{d}, Monday 4 November 2019 at the four prediction times. \textbf{e}, The weekday minus weekend contrast at the morning peak, Monday 4 November against Sunday 10 November 2019, on a diverging scale where red indicates a higher value on the weekday. The contrast is positive across essentially the whole network.}
    \label{fig:map}
\end{figure}

Within the twelve interval history window the gate is not uniform, and its shape follows the traffic conditions contained in the window rather than recency (Appendix~S5, Figure~\ref{fig:lag}). When predicting 09:44 on the Monday, the gate is lowest for the intervals between 05:14 and 06:44, at 0.349, and rises to its maximum of 0.456 at 09:14. When predicting 21:44 the maximum of 0.479 falls at 18:44, three hours before the prediction time, rather than at the most recent interval. The Sunday profile at 09:44 is close to flat, spanning 0.045 against the Monday range of 0.107. At 21:44 the Sunday profile is not flat, spanning 0.074, but it rises to a plateau of 0.425 rather than to the peak of 0.479 that the Monday profile reaches three hours before the prediction time. Mapping the gate onto the network shows the same pattern in space, with the whole network elevated at 21:44 and the weekday minus weekend contrast positive at 450 of the 498 segments at 09:44 (Figure~\ref{fig:map}).

\subsection{Relationship between the encoder and decoder GAT layers}

The remaining results concern the two graph attention layers and how they relate to one another. We first trace the order in which they give up their use of the graph as the penalty strengthens, and then remove each of them and retrain, so that the ordering the penalty produces can be compared against the consequence of not having the layer at all.

\begin{figure}[htbp]
    \centering
    \includegraphics[width=\textwidth]{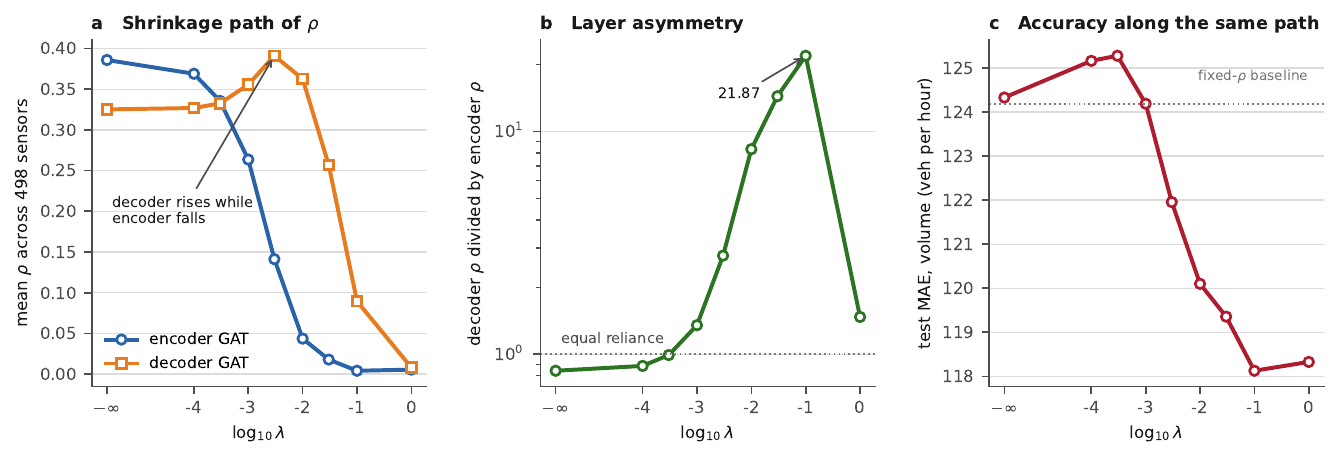}
    \caption{\textbf{Path traced by the two graph layers as the penalty strengthens.} \textbf{a}, Mean gate value in the encoder and decoder graph layers, averaged over the 498 sensors and the four prediction times on Sunday 10 November 2019. \textbf{b}, The ratio of the decoder mean to the encoder mean, on a logarithmic scale, with the value one marked. \textbf{c}, Test volume error along the same path, with the fixed-gate baseline marked.}
    \label{fig:shrinkage}
\end{figure}

We traced the mean gate value of each graph layer across the penalty range, using the four prediction times on Sunday 10 November 2019 (Figure~\ref{fig:shrinkage}). Without the penalty the encoder relies on its neighbours more than the decoder does, with mean gate values of 0.386 and 0.325. As the penalty strengthens the two layers diverge. The encoder mean falls monotonically to 0.004 at $\lambda = 0.1$. The decoder mean instead rises to a maximum of 0.391 at $\lambda = 0.003$ before falling to 0.089 at $\lambda = 0.1$. The ratio of decoder to encoder mean rises from 0.84 without the penalty to 21.87 at $\lambda = 0.1$. Both layers collapse at $\lambda = 1$, where the means are 0.005 and 0.008.

The point at which volume error is lowest coincides with the point at which the encoder gate is effectively closed. At $\lambda = 0.1$ the encoder mean gate is 0.004 and volume error is at its minimum of 118.1 vehicles per hour. Sensor-level gate values across the penalty range are shown in Supplementary Figure~S1.

\begin{figure}[htbp]
    \centering
    \includegraphics[width=\textwidth]{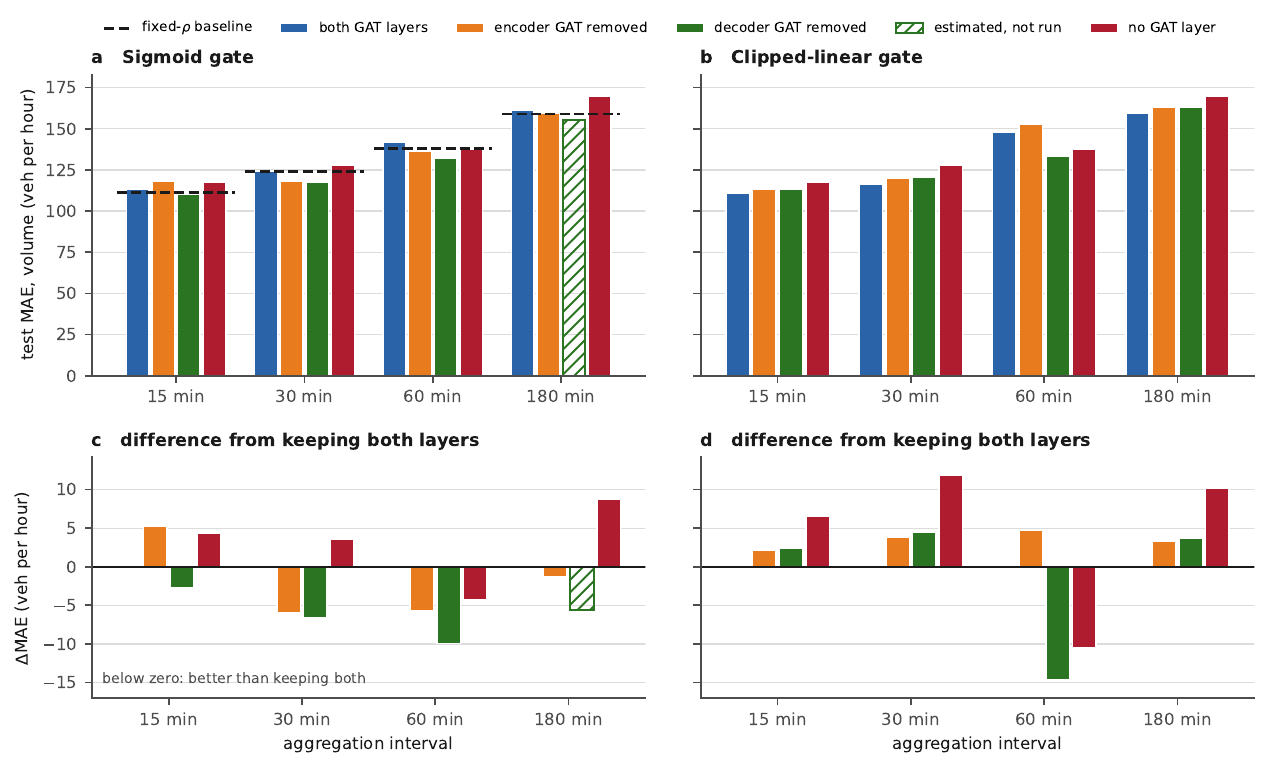}
    \caption{\textbf{Direct ablation of the two graph layers, under two gate nonlinearities.} \textbf{a} and \textbf{b}, Test volume error for four architectures at four aggregation intervals. The dashed line in \textbf{a} marks the fixed-gate ST-MetaNet baseline for that interval. \textbf{c} and \textbf{d}, The same values expressed as a difference from the architecture that keeps both graph layers, so that values below zero indicate a variant that is more accurate than keeping both. The decoder-removed variant at 180 minutes under the logistic gate was never run. It is shown hatched at a value estimated by applying the mean decoder-minus-encoder difference observed at the three intervals where both were run, and it is an estimate rather than a measurement.}
    \label{fig:ablation}
\end{figure}

We then removed each graph layer in turn and retrained, at four aggregation intervals and under two gate nonlinearities, to test whether that ordering is confirmed when the layers are actually taken away (Figure~\ref{fig:ablation}, Table~\ref{tab:ablation}).

Under the logistic gate, removing the decoder layer gives the lowest volume error at three of the four intervals. The values are 110.3 vehicles per hour at the 15 minute interval, 117.8 at 30 minutes and 131.8 at 60 minutes, against 118.3, 118.3 and 136.1 for removing the encoder layer, and 113.1, 124.3 and 141.7 for keeping both. At 180 minutes the decoder-removed variant was not run. Removing both layers is worse than removing either one at the 30, 60 and 180 minute intervals, at 127.8, 137.4 and 169.8 vehicles per hour.

\begin{table}[!ht]
\centering
\small
\caption{\textbf{Direct ablation of the two graph layers.} Test mean absolute error for volume, in vehicles per hour, on the held-out test period, for four architectures at four aggregation intervals and under the two gate nonlinearities. The lowest value in each column within each block is shown in bold. The model with no graph layer reduces to two stacked recurrent layers and is therefore identical under the two gate nonlinearities. The decoder-removed variant at the 180 minute interval under the logistic gate was not run.}
\label{tab:ablation}
\begin{tabular}{lrrrr}
\toprule
Architecture & 15 min & 30 min & 60 min & 180 min \\
\midrule
\multicolumn{5}{l}{\textit{Logistic gate}} \\
\hspace{1em}both graph layers & 113.12 & 124.33 & 141.74 & 160.95 \\
\hspace{1em}encoder graph layer removed & 118.32 & 118.32 & 136.07 & \textbf{159.62} \\
\hspace{1em}decoder graph layer removed & \textbf{110.35} & \textbf{117.77} & \textbf{131.77} & not run \\
\hspace{1em}no graph layer & 117.46 & 127.84 & 137.44 & 169.76 \\
\midrule
\multicolumn{5}{l}{\textit{Clipped linear gate}} \\
\hspace{1em}both graph layers & \textbf{110.97} & \textbf{116.03} & 147.99 & \textbf{159.59} \\
\hspace{1em}encoder graph layer removed & 113.13 & 119.91 & 152.67 & 162.88 \\
\hspace{1em}decoder graph layer removed & 113.40 & 120.46 & \textbf{133.37} & 163.22 \\
\hspace{1em}no graph layer & 117.46 & 127.84 & 137.44 & 169.76 \\
\bottomrule
\end{tabular}
\end{table}

\subsection{Sensitivity to the aggregation interval}

The whole penalty study was repeated at the 15, 60 and 180 minute aggregation intervals. The accuracy result is not general across them. An interior minimum below the fixed-gate baseline is specific to the 30 and 60 minute intervals, and at 15 and 180 minutes the unpenalised model is at least as accurate as any penalised one. The two findings the paper rests on are not affected. The divergence between the two graph layers under the penalty, and the gap between closing both gates and removing both layers, are present at every interval examined. The full response at each interval is given in Appendix~S3.

\section{Discussion}
\label{sec:discussion}

Internal quantities produced by trained networks are routinely read as measures of what the network depends on. Attention weights, gate values and gradient-based attributions are all used in this way, and in a forecasting setting they are seldom checked against anything outside the model. This study places one such measure inside the forecaster rather than beside it, so that its effect on accuracy is observable, and then checks what it reports against two references that exist outside the model. The measure proves informative about the traffic state at a given sensor and time, and regularising it provides a graded ablation whose conclusions can be tested by deleting the layers and retraining. What it does not settle is which of the two graph layers the architecture requires, and the reason it does not is instructive.

The first result is that the gate carries a signal about the physical system. It rises with the flow carried at a sensor, at Spearman correlations between 0.46 and 0.86 across the eight combinations of day and prediction time examined, while showing no consistent association with distance from London. It separates a weekday from the matched weekend day by an order of magnitude more at the morning and evening peaks than at 03:44, and not at all at 15:44. Within the history window it weights the intervals that contain the peak rather than the most recent intervals. These patterns are consistent with what has been reported using other explanation methods on other traffic networks. García-Sigüenza et al.~\citep{GarciaSiguenza2023} found that an adaptive graph convolutional recurrent network draws on simpler information at low flow and richer information as flow rises. Tygesen et al.~\citep{Unboxing} found that a learned interaction graph routes information towards congested locations regardless of their separation in space. The present result adds a direct, continuous and per-sensor association with a measured quantity, rather than a qualitative description of a mask. Part of this association is a property of the setting rather than a discovery about the model. Busy locations on a trunk road are more strongly coupled to their upstream and downstream neighbours than quiet ones, so a measure that tracks coupling would be expected to track flow. What the result establishes is that the gate does track it, at a magnitude that can be quoted, which is what allows the second test to be interpreted at all.

The penalty acts differently on the two forecast targets, and the difference is informative about what the graph layers carry. Constraining the gate lowers volume error and raises speed error, and the separation widens as the penalty strengthens. Volume at a sensor is close to a conserved quantity along a corridor, so what enters a segment is strongly determined by what left the segments upstream of it, and a neighbour aggregate is a direct estimate of that. Speed is not conserved in the same way. It is bounded above by the speed limit, varies over a narrow range for most of the day, and is determined more by the local density at the sensor than by conditions one junction away. A constraint that removes graph information should therefore cost more for speed than for volume, which is what the measurements show. That the volume error also falls rather than merely holding steady indicates that the unconstrained gate was using more neighbour information than the volume task required. These results are single-run differences of about 6 vehicles per hour and 0.01 miles per hour, so the direction is better supported than the magnitude.

The second result is that the same measure does not identify which graph layer the architecture needs, and it fails in two distinct ways. The first is that the ordering it produces cannot be checked. Under the penalty the encoder layer gives up its reliance first, falling from a mean gate value of 0.386 to 0.004 while the decoder layer rises to a maximum of 0.391 before falling to 0.089. Read as an importance ranking this favours the decoder layer, but the ablation that would confirm or refute it separates the two variants by 0.55 vehicles per hour at the 30 minute interval, against a typical spread of 4.2 across the final validation checkpoints of a single run. The per-configuration figures behind that spread are given in Appendix~S8. The ranking is not corroborated, and with these models it cannot be.

The second failure is sharper, because it does not depend on distinguishing the two layers. Closing a gate is not the same operation as removing the layer it controls. For one layer the two coincide, and the penalised model with the encoder gate at 0.004 lands within 0.19 vehicles per hour of the model with the encoder layer removed and retrained. For both layers they do not, and the penalised model with both gates below 0.01 is 9.51 vehicles per hour more accurate than the model with both layers removed and retrained. A gate at 0.004 leaves the layer in the computational graph, so the parameters upstream of it were shaped by twelve intervals of graph exchange during the part of training before the gate closed, and the state entering the recurrent layer below it is not the state a model trained without the layer would have produced. The internal reading is therefore locally faithful and globally not, and the gap is seventeen times the difference the ranking rests on.

Two features of the setting together account for the disagreement. The first is that the two graph layers can substitute for one another over part of the range. Under the logistic gate, at the 30, 60 and 180 minute intervals, removing either layer alone matches or improves on keeping both while removing both is worse than removing either. At 15 minutes this does not hold. There, keeping both layers gives 113.1 vehicles per hour against 118.3 for removing the encoder layer, and removing both layers gives 117.5, which is better than removing the encoder layer alone. Where substitution is available, a single opportunity to exchange information across the network appears to be sufficient, and providing two is not only unnecessary but slightly harmful. When two components can substitute for one another, a penalty applied to both can be satisfied by closing either, and the optimiser will close whichever is cheaper to close.

The second feature determines which one is cheaper, and it is a property of the penalty rather than of the architecture. The penalty is a mean over gate values, and the encoder gate is evaluated at twelve intervals against the decoder gate's one, so twelve of every thirteen penalised values belong to the encoder. The encoder gate network therefore receives about twelve times the aggregate penalty pressure. Position in the computational graph plausibly contributes as well, since the encoder layer sits twelve recurrent steps from the loss while the decoder layer sits one step away, so the task gradient reaching the encoder gate is attenuated relative to the penalty gradient acting on it. Both influences push in the same direction, and neither is a statement about how much either layer contributes to the forecast. The penalty path therefore records which pathway is cheapest to switch off under a particular penalty, while the architecture is held fixed. Ablation asks a different question, namely what the architecture can achieve after being retrained without a pathway. These two questions coincide only when the components are not substitutable and the diagnostic does not itself favour one of them. This distinction has been documented in image classification, where measures of unit importance computed on a trained network have been shown to disagree with the effect of removing the unit \citep{SingleDirections}, and where retraining after removal changes the ranking that attribution methods produce \citep{ROAR}. The result reported here indicates that spatio-temporal traffic forecasters are subject to the same limitation.

The result generalises beyond the specific gate studied. Any diagnostic computed from the internal state of a trained model, without retraining, observes a single configuration of that model. Such a diagnostic can identify a component that is currently inactive, but it cannot separate a component that is unnecessary from one that is redundant with another. Attention weights, gate magnitudes and shrinkage paths all share this property. The practical implication for work that reports attention patterns as evidence of architectural importance is that the ablation is not an optional confirmation. It is the measurement, and the internal quantity is a hypothesis about it.

The two readings are also separated by cost, which sharpens the practical stake. Removing the encoder graph layer saves 65.5 minutes of the 121.0 minute training time at the 30 minute interval, and removing the decoder layer saves 12.5 minutes. A practitioner who pruned on the basis of the penalty path would have removed the encoder layer, and would have obtained a model that trains in less than half the time and is no less accurate at three of the four intervals. The outcome would have been acceptable. It would not have been supported by the evidence, because the ablation cannot separate the two single-layer variants at all. They differ by 0.55 vehicles per hour against a typical spread of 4.2 across the final checkpoints of one run, so the ranking the penalty path implies is neither confirmed nor refuted. What the ablation does establish is that the two layers are largely redundant with one another, which is why removing either was survivable and why removing both was not. A decision that lands well because two components substitute for one another is not evidence that the diagnostic which prompted it identified the right one, and at the 15 minute interval the same decision costs 5.2 vehicles per hour against keeping both layers.

The design that produced this comparison is its main methodological strength. Traffic forecasting supplies a measured external quantity at every node and every timestep, so an internal reliance measure can be checked directly rather than assessed for plausibility. It also permits the two graph layers to be removed and the model retrained at modest cost, so the internal ranking has a direct counterfactual to be tested against. Few settings in which explanation methods are applied offer both.

\subsection{Limitations}
\label{subsec:limitations}

Five limitations bound what these results support.

First, the models are not trained to convergence and no model selection is performed. Training runs for a fixed number of updates at a constant learning rate, without early stopping, and the parameters at the final update are the ones evaluated. Validation error is still oscillating at termination, with a standard deviation of between 4.00 and 8.49 vehicles per hour across the final five evaluations and a final value that is the lowest recorded in only one of six configurations. Every configuration was also trained once, from a single random seed. The pipeline is deterministic and repeated execution returns identical results, so what is unquantified is the variation a different initialisation would produce, which the checkpoint spread bounds from below rather than measures. The consequence is that no comparison smaller than about 4 vehicles per hour is interpretable, which is why the ordering between the two single-layer variants is reported here as untestable rather than as resolved. The two contrasts the argument rests on, 6.0 to 6.6 vehicles per hour for removing a layer and 9.5 for closing both gates against removing both layers, sit above that threshold but not far above it. Training to convergence with best-checkpoint selection, across several seeds, is what this study most needs and is what would allow the ordering question to be answered rather than set aside.

Second, the ablation outcome is not stable across the gate nonlinearity. Replacing the logistic gate with a clipped linear gate reverses the finding that either layer can be removed, and under the clipped linear gate keeping both layers is best at three of the four intervals. Those differences are 2.2 to 4.5 vehicles per hour and are at or below the stability threshold above, so the instability is indicative rather than established. It points the same way as the main argument, because a diagnostic whose architectural conclusion moves with the choice of squashing function is not identifying a property of the architecture. Both gate variants are reported here for that reason.

Third, the shrinkage path was computed from four prediction times on a single Sunday, because the corresponding weekday arrays at multiple penalty strengths were not retained. Given that the first result of this study is that the gate differs between a weekday and a weekend day, a path computed on a weekday could differ in magnitude. The direction of the divergence between the two layers is unlikely to reverse, because it is driven by the construction of the penalty rather than by the traffic state, but this has not been shown. Regenerating the weekday arrays from the stored model weights would settle it.

Fourth, the study covers one network, one year and one model family. The 498 sensors span the strategic road network of England, which is a single national system with a particular topology and a particular degree of congestion. Whether the disagreement between an internal reliance measure and direct ablation appears in models built on urban arterial networks, or in architectures that place their graph layers differently, is not established by this design.

Fifth, and most fundamentally, the study contains no comparison against an established explanation method. GNNExplainer, PGExplainer and GraphMask are the standard tools for this task, and none is run here. These methods address the importance of individual edges rather than the aggregate reliance of a node update, so the comparison is not direct, but its absence limits what can be said about the relative merits of the approach. The ideal version of this study would apply an edge masking method to the same network and the same trained models, subject each of the two readouts to the same ablation check, and report whether both fail in the same way or only one does. That comparison would establish whether the limitation demonstrated here is a property of gate-based reliance measures specifically or of post-hoc importance measures in general.

\section{Conclusion}
\label{sec:conclusion}

This study introduced a gated graph attention layer for short-term traffic forecasting and applied it to ST-MetaNet on 498 loop detectors of the strategic road network of England. Because the gate is trained with the model rather than fitted to it afterwards, its effect on forecast accuracy is observable across the whole regularisation range, and mild regularisation gives a modest improvement over both the unpenalised and the fixed-gate model at the intermediate aggregation intervals. The gate provides a continuous account of where and when the model draws on the road network. It rises with the volume a sensor carries, at Spearman correlations of 0.46 to 0.86, and it separates a weekday from the matched Sunday at the peak hours, so the spatial information flow it reports is consistent with the traffic the same sensors record. Regularising the gate further provides a graded ablation, and comparing that against deleting the layers and retraining shows the two graph attention layers to be largely redundant with one another. Removing either alone leaves accuracy at least as good as keeping both, while removing both degrades it substantially. The same comparison marks the limit of what an internal reading can settle. Closing one gate reproduces deleting that layer, closing both does not reproduce deleting both, and the order in which the two gates close under the penalty is not confirmed by removing them. A measure of this kind therefore describes how a trained model uses its graph, and a decision to simplify the architecture should rest on retraining without the component in question.

\section*{Acknowledgements}

This research was supported by the Cambridge Commonwealth, European and International Trust. Additional support was provided by the Martin Centre for Architectural and Urban Studies. The authors gratefully acknowledge National Highways for providing the traffic data used in this study. The authors also thank colleagues and external experts for valuable discussions and feedback. The views expressed are those of the authors and do not necessarily reflect those of the supporting institutions.

\section*{Funding}

This research did not receive any specific grant from funding agencies in the public, commercial, or not-for-profit sectors. The supporting institutions named above had no role in study design, in the collection, analysis and interpretation of data, in the writing of the report, or in the decision to submit the article for publication.

\section*{Declaration of competing interests}

The authors declare that they have no known competing financial interests or personal relationships that could have appeared to influence the work reported in this paper.

\section*{Data availability}

All traffic data used in this study are publicly available from National Highways through the Traffic Information System API \citep{TRIS2024}, under the Open Government Licence. The processed 2019 sensor series, the network definition, and the trained model weights are available in the study repository.

\section*{Code availability}

The model code, the trained weights for every penalty strength, and the training logs are available at \url{https://github.com/yueli901/Gated-GAT}, with documentation of the system configuration, the software dependencies, the data files and the unit conversions the logs require. The repository README states what the tree does and does not reproduce. In particular, the four ablation variants of Table~\ref{tab:ablation} were produced by editing the source in place rather than through a configuration option, so they cannot be regenerated from the released tree without that edit.

\bibliographystyle{elsarticle-num}
\bibliography{gatreg}

\clearpage

\setcounter{section}{0}
\setcounter{figure}{0}
\setcounter{table}{0}
\setcounter{equation}{0}
\renewcommand{\thesection}{S\arabic{section}}
\renewcommand{\thesubsection}{S\arabic{section}.\arabic{subsection}}
\renewcommand{\thetable}{S\arabic{table}}
\renewcommand{\thefigure}{S\arabic{figure}}
\renewcommand{\theequation}{S\arabic{equation}}

\begin{center}
{\large\bfseries Supplementary Material}\\[6pt]
{\large Explaining spatial information flow in short-term traffic forecasting models\\
using a gated graph attention network}
\end{center}

\vspace{1.5em}

\section*{Contents}

\noindent
\begin{tabular}{@{}p{0.115\textwidth}p{0.70\textwidth}r@{}}
\toprule
Item & Content & Page \\
\midrule
Appendix S1 & Sensor-level gate values across the penalty range & \pageref{sec:s1} \\
Appendix S2 & Bootstrap distributions for the accuracy comparisons & \pageref{sec:s2} \\
Appendix S3 & Behaviour across the four aggregation intervals & \pageref{sec:s3} \\
Appendix S4 & Complete inventory of trained configurations & \pageref{sec:s4} \\
Appendix S5 & Gate values across the history window, all prediction times & \pageref{sec:s5} \\
Appendix S6 & Correlations between the gate and the measured traffic state & \pageref{sec:s6} \\
Appendix S7 & Implementation notes and departures from the original ST-MetaNet & \pageref{sec:s7} \\
Appendix S8 & Training stability and the interpretable effect size & \pageref{sec:s8} \\
Appendix S9 & Provenance of the clipped-linear gate runs & \pageref{sec:s9} \\
\bottomrule
\end{tabular}

\clearpage

\section{Sensor-level gate values across the penalty range}
\label{sec:s1}

\begin{figure}[htbp]
    \centering
    \includegraphics[width=0.88\textwidth]{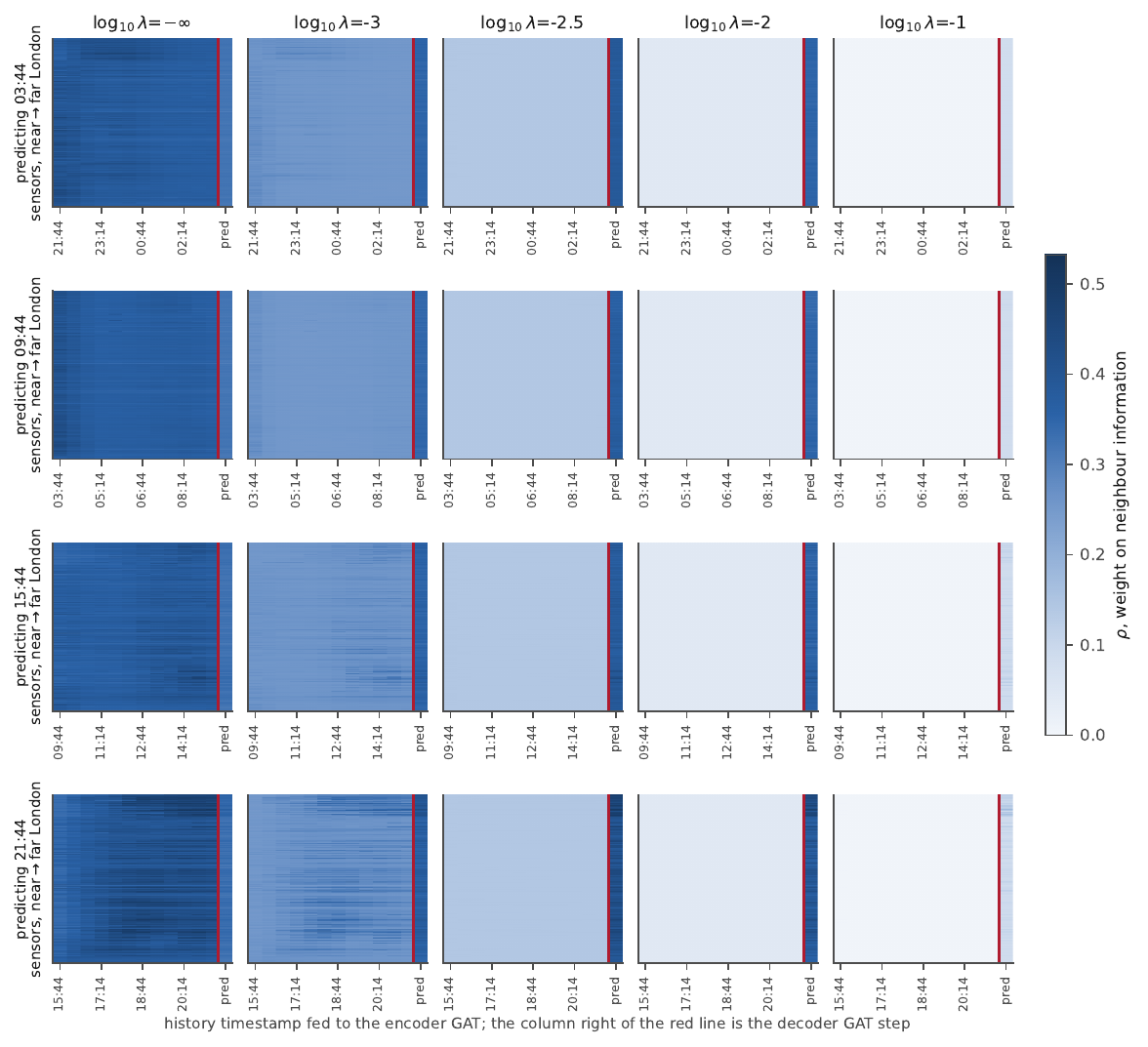}
    \caption{\textbf{Gate values for every sensor across the penalty range.} Rows are the four prediction times and columns are five penalty strengths. Within each panel, each row of the image is one of the 498 sensors, sorted by distance from central London, and each column is one interval of the input window. The twelve columns to the left of the red rule are the encoder intervals and the single column to its right is the decoder interval. Colour gives the gate value on a common scale across all panels. Values are from Sunday 10 November 2019.}
    \label{fig:s1}
\end{figure}

Section 3.3 of the main text reports the mean gate value of each graph layer at each penalty strength. Figure~\ref{fig:s1} shows the underlying values for every sensor and every interval in the history window, so that the collapse of the encoder layer and the delayed collapse of the decoder layer can be seen at the level of individual sensors rather than only in the mean.

Two features are visible that the means do not carry. The encoder collapse is close to uniform across sensors, with no group of sensors retaining reliance while the rest give it up. The decoder column, immediately to the right of the vertical rule, remains dark at penalty strengths where the encoder columns have faded to near white, which is the sensor-level form of the divergence reported in the main text.

The gate values in this figure were produced on Sunday 10 November 2019. The matched Monday values at multiple penalty strengths were not retained in the working tree and would require the saved model weights to be re-run.

\clearpage

\section{Bootstrap distributions for the accuracy comparisons}
\label{sec:s2}

\begin{figure}[htbp]
    \centering
    \includegraphics[width=\textwidth]{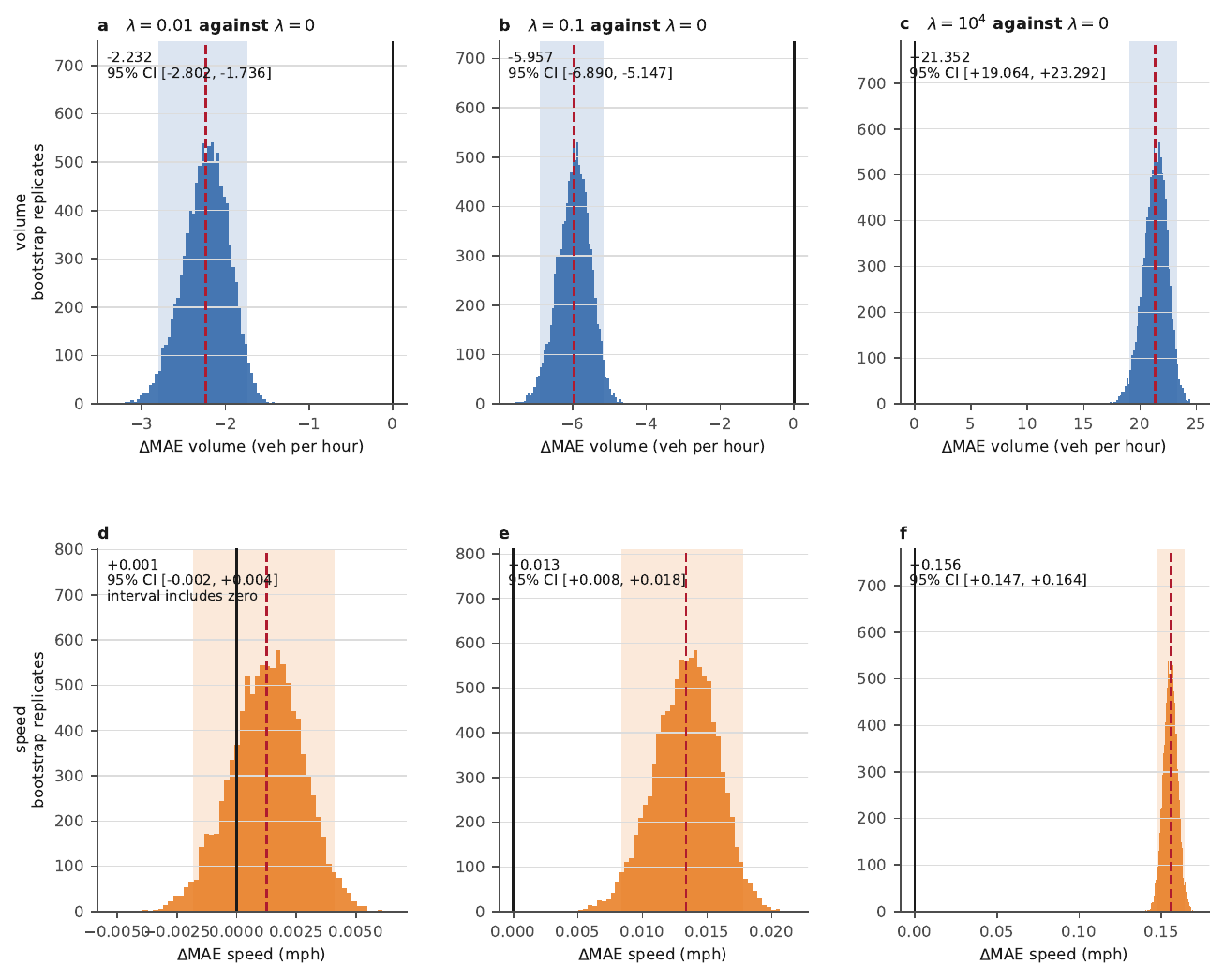}
    \caption{\textbf{Bootstrap replicate distributions for the accuracy comparisons.} Each panel shows 10{,}000 paired replicates of the difference in mean absolute error between two models, resampled over the 73 test days with replacement. \textbf{a} to \textbf{c}, Volume, in vehicles per hour. \textbf{d} to \textbf{f}, Speed, in miles per hour. The dashed red line is the observed difference on the full test period, the solid black line is zero, and the shaded band is the 95\% interval. Negative values indicate lower error under the penalised model. Missing observations are excluded using the recorded indicator channel rather than a value threshold.}
    \label{fig:s2}
\end{figure}

Section 3.1 of the main text reports the paired day-block bootstrap comparison between penalised and unpenalised models. Figure~\ref{fig:s2} shows the replicate distributions behind those intervals.

The procedure resamples whole test days with replacement, 10{,}000 times, and recomputes the difference in mean absolute error between the two models on each resample. Both models are evaluated on exactly the same cells, so the comparison is paired at the level of the individual observation. Missing observations are excluded using the recorded indicator channel rather than a value threshold. This matters for speed, because a missing speed value is stored as a positive constant after inverse standardisation and would pass a positivity filter, contaminating about one tenth of the cells.

The volume differences at $\lambda = 0.01$ and $\lambda = 0.1$ and the large deterioration at $\lambda = 10^{4}$ all have intervals well clear of zero. The speed difference at $\lambda = 0.01$ has an interval that includes zero. The speed difference at $\lambda = 0.1$ does not, and is positive, meaning the penalty is associated with slightly worse speed forecasts at that strength.

\clearpage

\section{Behaviour across the four aggregation intervals}
\label{sec:s3}

\begin{figure}[htbp]
    \centering
    \includegraphics[width=0.98\textwidth]{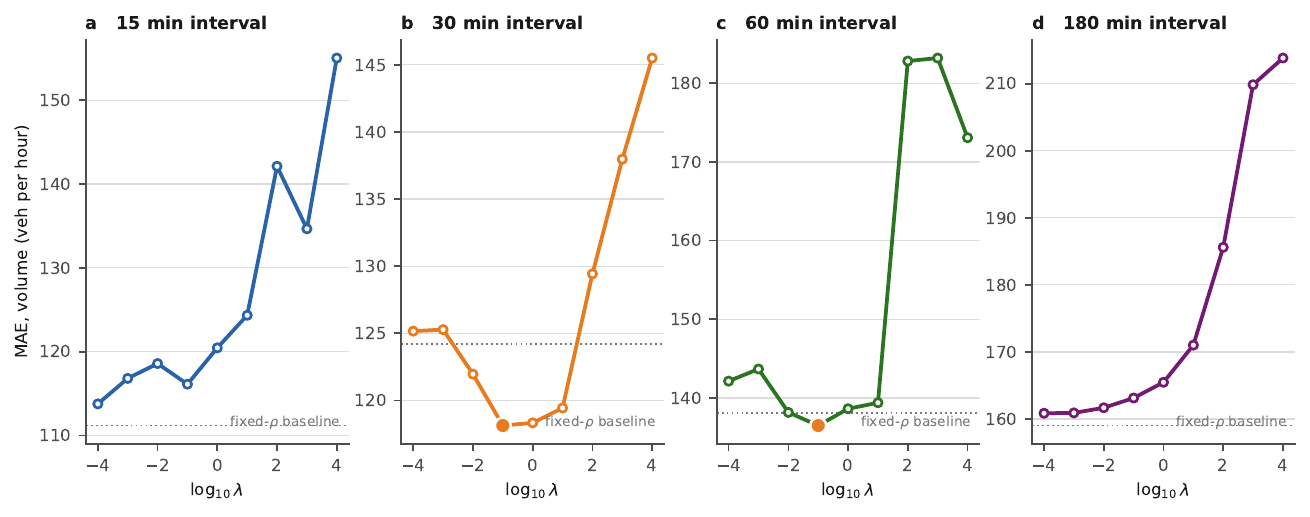}
    \caption{\textbf{Volume error against penalty strength at each aggregation interval.} Each panel gives the test mean absolute error for volume in vehicles per hour at one aggregation interval, against the fixed-gate ST-MetaNet baseline for that interval, shown as a dotted line. An orange marker indicates a minimum that lies below the baseline. Volume is expressed as a flow rate throughout, so the four panels are on a common unit.}
    \label{fig:s3}
\end{figure}

The main text reports the penalty study at the 30 minute aggregation interval and states in Section 3.4 that the interior minimum in volume error is specific to the 30 and 60 minute intervals. This appendix gives the full response at each of the four intervals. Figure~\ref{fig:s3} expresses each against the fixed-gate baseline for its own interval so that the four panels are comparable.

At the 15 minute interval the unpenalised model gives the lowest error of the penalised set. At 180 minutes it is beaten only marginally, by 0.1 vehicles per hour at $\lambda = 10^{-4}$. At 30 minutes a minimum falls at $\lambda = 0.1$ and lies below the baseline. At 60 minutes a minimum also falls at $\lambda = 0.1$, but the curve is not monotone on either side of it and the improvement over the baseline is smaller. None of the four curves is monotone in the penalty. Error is much higher at the top of the range than at the bottom at every interval, but it does not rise at every step: at 15 minutes it falls from 142.11 to 134.65 vehicles per hour between $\lambda = 10^{2}$ and $\lambda = 10^{3}$, and at 60 minutes from 183.20 to 173.08 between $\lambda = 10^{3}$ and $\lambda = 10^{4}$.

The accuracy result reported in the main text is therefore specific to the intermediate aggregation intervals. The two findings the paper rests on, namely the association between the gate and the measured traffic state and the disagreement between the penalty path and the ablation, do not depend on it. Both are present at every interval examined.

\clearpage

\section{Complete inventory of trained configurations}
\label{sec:s4}

Table~\ref{tab:s4} lists every configuration reported in the main text with its aggregation interval, gate nonlinearity, penalty strength and architecture. All configurations use the same data, the same split and the same training protocol, and differ only in the factor named. The number of passes over the training set is set per interval so that the number of gradient updates is held approximately constant at 767.

\begin{table}[htbp]
\centering
\small
\caption{\textbf{Configurations trained for this study.} Volume errors are in vehicles per hour and speed errors in miles per hour, on the held-out test period. Passes gives the number of complete passes over the training set.}
\label{tab:s4}
\begin{tabular}{llrlrrrr}
\toprule
Interval & Gate & $\lambda$ & Architecture & Passes & MAE vol. & MAE spd. & RMSE vol. \\
\midrule
15 min & fixed & -- & both layers & 1 & 111.19 & 1.608 & 187.21 \\
15 min & logistic & 0 & both layers & 1 & 113.12 & 1.600 & 189.36 \\
15 min & logistic & 0 & encoder removed & 1 & 118.32 & 1.589 & 194.54 \\
15 min & logistic & 0 & decoder removed & 1 & 110.35 & 1.609 & 186.38 \\
15 min & logistic & 0 & no graph layer & 1 & 117.46 & 1.612 & 197.56 \\
15 min & clipped & 0 & both layers & 1 & 110.97 & 1.616 & 188.54 \\
15 min & clipped & 0 & encoder removed & 1 & 113.13 & 1.608 & 191.11 \\
15 min & clipped & 0 & decoder removed & 1 & 113.40 & 1.615 & 188.82 \\
\midrule
30 min & fixed & -- & both layers & 2 & 124.19 & 1.586 & 207.00 \\
30 min & logistic & 0 & both layers & 2 & 124.33 & 1.550 & 199.23 \\
30 min & logistic & 0 & encoder removed & 2 & 118.32 & 1.565 & 202.01 \\
30 min & logistic & 0 & decoder removed & 2 & 117.77 & 1.571 & 199.14 \\
30 min & logistic & 0 & no graph layer & 2 & 127.84 & 1.578 & 213.15 \\
30 min & logistic & 0.1 & both layers & 2 & 118.13 & 1.564 & 199.05 \\
30 min & clipped & $10^{-3}$ & both layers & 2 & 115.42 & 1.557 & 196.50 \\
30 min & clipped & 0 & both layers & 2 & 116.03 & 1.555 & 195.62 \\
30 min & clipped & 0 & encoder removed & 2 & 119.91 & 1.577 & 203.97 \\
30 min & clipped & 0 & decoder removed & 2 & 120.46 & 1.569 & 202.47 \\
\midrule
60 min & fixed & -- & both layers & 4 & 138.13 & 1.667 & 224.25 \\
60 min & logistic & 0 & both layers & 4 & 141.74 & 1.674 & 226.15 \\
60 min & logistic & 0 & encoder removed & 4 & 136.07 & 1.715 & 224.93 \\
60 min & logistic & 0 & decoder removed & 4 & 131.77 & 1.657 & 218.78 \\
60 min & logistic & 0 & no graph layer & 4 & 137.44 & 1.684 & 226.54 \\
60 min & clipped & 0 & both layers & 4 & 147.99 & 1.695 & 230.54 \\
60 min & clipped & 0 & encoder removed & 4 & 152.67 & 1.664 & 233.66 \\
60 min & clipped & 0 & decoder removed & 4 & 133.37 & 1.646 & 223.11 \\
\midrule
180 min & fixed & -- & both layers & 12 & 159.02 & 1.872 & 257.17 \\
180 min & logistic & 0 & both layers & 12 & 160.95 & 1.825 & 260.26 \\
180 min & logistic & 0 & encoder removed & 12 & 159.62 & 1.861 & 265.21 \\
180 min & logistic & 0 & no graph layer & 12 & 169.76 & 1.954 & 290.11 \\
180 min & clipped & 0 & both layers & 12 & 159.59 & 1.852 & 257.30 \\
180 min & clipped & 0 & encoder removed & 12 & 162.88 & 1.842 & 269.09 \\
180 min & clipped & 0 & decoder removed & 12 & 163.22 & 1.995 & 268.02 \\
\bottomrule
\end{tabular}
\end{table}

\clearpage

\section{Gate values across the history window}
\label{sec:s5}

Section 3.2 of the main text reports the shape of the gate across the twelve intervals of the history window at two prediction times. Table~\ref{tab:s5} gives the values at all four prediction times on both days, so that the flat profiles can be compared against the peaked ones directly.

The Monday profile at 09:44 falls from 0.428 at the oldest interval to a minimum of 0.349 in the early morning and then rises to 0.456 at the most recent interval, a range of 0.107. The matched Sunday profile spans 0.044. At 21:44 the Monday maximum of 0.479 falls at 18:44, three hours before the prediction time, whereas the Sunday profile reaches 0.425 and holds it as a plateau. At 03:44 and 15:44 both days are close to flat.

\begin{table}[htbp]
\centering
\small
\caption{\textbf{Mean gate value at each interval of the history window.} Values are means across the 498 sensors under the unpenalised model. Columns give the twelve encoder intervals in chronological order, oldest first, followed by the single decoder interval. Monday is 4 November 2019 and Sunday is 10 November 2019.}
\label{tab:s5}
\begin{tabular}{llrrrrrrrrrrrrr}
\toprule
Day & Predicting & \multicolumn{12}{c}{Encoder intervals, oldest to most recent} & Dec. \\
\cmidrule(lr){3-14}
 & & 1 & 2 & 3 & 4 & 5 & 6 & 7 & 8 & 9 & 10 & 11 & 12 & \\
\midrule
Mon & 03:44 & .396 & .404 & .399 & .390 & .381 & .381 & .380 & .379 & .377 & .377 & .376 & .376 & .320 \\
Mon & 09:44 & .428 & .392 & .359 & .350 & .349 & .349 & .351 & .363 & .392 & .442 & .455 & .456 & .337 \\
Mon & 15:44 & .347 & .376 & .386 & .390 & .386 & .380 & .380 & .379 & .380 & .377 & .378 & .377 & .323 \\
Mon & 21:44 & .342 & .359 & .384 & .401 & .424 & .456 & .479 & .474 & .452 & .425 & .410 & .400 & .328 \\
\midrule
Sun & 03:44 & .402 & .400 & .392 & .387 & .388 & .390 & .384 & .377 & .374 & .371 & .369 & .370 & .321 \\
Sun & 09:44 & .411 & .393 & .375 & .366 & .367 & .368 & .371 & .374 & .374 & .375 & .372 & .367 & .319 \\
Sun & 15:44 & .369 & .370 & .366 & .367 & .367 & .371 & .380 & .385 & .388 & .393 & .391 & .384 & .327 \\
Sun & 21:44 & .351 & .367 & .386 & .399 & .414 & .422 & .425 & .420 & .421 & .424 & .424 & .419 & .334 \\
\bottomrule
\end{tabular}
\end{table}

\clearpage

\section{Correlations between the gate and the measured traffic state}
\label{sec:s6}

Table~\ref{tab:s6} gives the full set of correlations summarised in Section 3.2 of the main text, including the comparison against distance from central London.

\begin{table}[htbp]
\centering
\small
\caption{\textbf{Spearman rank correlations between the mean encoder gate and two node-level quantities.} Correlations are across sensors at a single prediction time under the unpenalised model. Sensors recording fewer than 40 vehicles per hour are excluded, which is why $n$ varies. Distance is the Euclidean distance in degrees from each sensor to central London.}
\label{tab:s6}
\begin{tabular}{llrrrrr}
\toprule
Day & Predicting & $n$ & $r$ (flow) & $P$ (flow) & $r$ (distance) & $P$ (distance) \\
\midrule
Monday 4 Nov & 03:44 & 402 & $+0.588$ & $<.001$ & $-0.083$ & .10 \\
Monday 4 Nov & 09:44 & 453 & $+0.461$ & $<.001$ & $-0.027$ & .56 \\
Monday 4 Nov & 15:44 & 451 & $+0.728$ & $<.001$ & $+0.114$ & .02 \\
Monday 4 Nov & 21:44 & 442 & $+0.856$ & $<.001$ & $-0.122$ & .01 \\
\midrule
Sunday 10 Nov & 03:44 & 344 & $+0.710$ & $<.001$ & $-0.120$ & .03 \\
Sunday 10 Nov & 09:44 & 453 & $+0.519$ & $<.001$ & $+0.097$ & .04 \\
Sunday 10 Nov & 15:44 & 453 & $+0.579$ & $<.001$ & $+0.192$ & $<.001$ \\
Sunday 10 Nov & 21:44 & 449 & $+0.849$ & $<.001$ & $-0.066$ & .16 \\
\bottomrule
\end{tabular}
\end{table}

The association with flow holds at every combination of day and prediction time, with correlations between 0.46 and 0.86. The association with distance from London changes sign four times across the eight combinations and reaches at most 0.19 in absolute value.

\clearpage

\section{Implementation notes and departures from the original ST-MetaNet}
\label{sec:s7}

\subsection*{S7.1 The node update}

The original ST-MetaNet combines the node state and the neighbour aggregate multiplicatively, applying a learned scalar to the aggregate. The implementation used here forms a convex combination instead, so that the weight on the neighbour aggregate and the weight on the node's own state sum to one. This is what allows the gate to be read as a share of the update, and it is the reason the quantity is bounded in the unit interval. The fixed-gate baseline in the main text uses the same convex combination with a single learned scalar for each graph, so the comparison between the baseline and the gated model isolates the effect of letting the weight vary by sensor and by timestep.

\subsection*{S7.2 Where the penalty is applied}

The penalty is applied to the gate values, which are an intermediate output of the model, rather than to the parameters of the network that produces them. Applying it to the parameters would leave the mapping from parameters to gate values free to compensate. The penalty is the mean absolute gate value over all sensors, all intervals, both dual graphs and both the encoder and the decoder. Because the gate is bounded in the unit interval, the absolute value is the value itself.

\subsection*{S7.3 What the penalty cannot do}

The logistic gate cannot reach exactly zero, so the penalty compresses gate values towards zero without setting any of them to it. The method therefore gives a continuous path rather than the discrete inclusion decision that an edge masking method produces. The clipped linear gate reported alongside it admits exact zeros. Both are reported because the ablation outcome differs between them, which is one of the findings of the study.

\subsection*{S7.4 Training details not carried in the main text}

Model parameters are initialised from the default TensorFlow initialisers with all random seeds fixed at 2. Gradients are clipped elementwise at $\pm 5$ rather than by global norm. The learning rate is held constant at $0.01$ throughout training, with no decay and no early stopping. Scheduled sampling is present in the implementation but is inactive at the output length of one interval used here, because the decoder loop executes a single step and takes the branch that uses the encoder state rather than a previous prediction.

\subsection*{S7.5 Units of the recorded metrics}

The training code records volume as the mean of the constituent 15 minute counts and speed as the sum of the constituent 15 minute mean speeds. Both are converted before reporting. Volume is multiplied by four to give a flow rate in vehicles per hour, which is valid at every aggregation interval because the recorded quantity is already a per-15-minute mean. Speed is divided by the number of 15 minute periods in the aggregation interval, which is one, two, four and twelve at the 15, 30, 60 and 180 minute intervals respectively. Both conversions are exact on cells with no missing constituent measurement, and cells with a missing constituent measurement are excluded from every metric.

\subsection*{S7.6 Reported metric definition}

The reported error is the mean of per-batch means rather than a mean pooled over all observations. Because the final batch of each split contains fewer samples than the others, this weights the last samples of a split slightly more than a pooled mean would. Recomputing the volume error for the unpenalised model as a pooled mean over the saved test predictions gives 124.64 vehicles per hour against the 124.33 obtained as a mean of batch means. The difference is below 0.5 vehicles per hour, is in the same direction for every configuration, and does not affect any comparison reported in this study.

\clearpage

\section{Training stability and the interpretable effect size}
\label{sec:s8}

Training runs for a fixed number of updates at a constant learning rate. There is no early stopping and no best-checkpoint selection, so the parameters evaluated are those at the final update. Validation error was recorded at seventeen points during each run. Table~\ref{tab:s8} gives the last five recorded values for the six configurations at the 30 minute interval, together with the final and the best value of the whole run.

Validation error is still oscillating at termination in every configuration. The standard deviation of the final five evaluations is between 4.00 and 8.49 vehicles per hour with a median of 4.23. In five of the six configurations the final evaluation is not the lowest of the run, and in the sixth, the decoder-removed variant, it is the lowest by less than 0.01 vehicles per hour. Any comparison between configurations that is smaller than about 4 vehicles per hour is therefore not interpretable from a single run. This is the threshold used in Section 3.3 of the main text to set aside the ordering between the two single-layer variants, whose difference is 0.55 vehicles per hour, while retaining the contrasts of 6.0, 6.6 and 9.5 vehicles per hour.

\begin{table}[htbp]
\centering
\small
\caption{\textbf{Validation volume error over the final five evaluations, 30 minute interval.} Values are in vehicles per hour. Best gives the lowest value recorded at any point in the run, and Final gives the value at the last update, which is the parameter set that produced the reported test metrics.}
\label{tab:s8}
\begin{tabular}{lrrrrrrrr}
\toprule
Configuration & \multicolumn{5}{c}{Last five evaluations} & SD & Final & Best \\
\midrule
ST-MetaNet, fixed gate & 124.5 & 123.8 & 143.9 & 124.8 & 129.3 & 8.49 & 129.33 & 123.81 \\
both graph layers & 129.2 & 119.7 & 125.1 & 130.4 & 126.9 & 4.21 & 126.93 & 119.66 \\
encoder layer removed & 122.5 & 121.8 & 132.2 & 126.1 & 123.4 & 4.25 & 123.42 & 121.78 \\
decoder layer removed & 129.1 & 123.4 & 139.6 & 123.4 & 123.2 & 7.06 & 123.19 & 123.19 \\
no graph layer & 129.4 & 125.8 & 135.8 & 129.8 & 134.4 & 4.05 & 134.40 & 125.79 \\
$\lambda = 0.1$ & 126.9 & 120.6 & 129.8 & 129.0 & 122.8 & 4.00 & 122.76 & 120.57 \\
\bottomrule
\end{tabular}
\end{table}

\section{Provenance of the clipped-linear gate runs}
\label{sec:s9}

Two versions of the clipped-linear gate exist in the working tree. The version used for every configuration reported in this study is
\[
\rho = \min(\max(x,0),1),
\]
which is bounded in the unit interval and preserves the reading of $\rho$ as a share of the node update. A second version,
\[
\rho = \begin{cases} 0 & x \le -0.5 \\ x & -0.5 < x \le 0.5 \\ 1 & x > 0.5, \end{cases}
\]
was written later. It is discontinuous at both breakpoints and admits negative values, so under it the node update is no longer a convex combination. It was used for one run, which is not reported here.

The distinction is recoverable from timestamps. The source file carrying the second version was last modified on 14 April 2024 at 10:05. Every clipped-gate configuration in Table~\ref{tab:s4} began training before that time, the latest at 10:04 on the same day, and the 30 minute unpenalised clipped run reproduces bit for bit a run executed on 11 April in a separate directory whose source carries the first version. The single run that began after the modification, at 19:49 on 14 April, gives a volume error of 120.0 vehicles per hour and is excluded from all tables and figures.

\end{document}